\documentclass[prodmode,acmtecs]{acmsmall}

\usepackage[ruled]{algorithm2e}

\SetAlFnt{\normal}
\SetAlCapFnt{\small}
\SetAlCapNameFnt{\small}
\SetAlCapHSkip{0pt}
\IncMargin{-\parindent}

\graphicspath{{./figs/}{./fig/}{./newfigs/}{./rev_figs/}}

\usepackage[bf]{caption}

\usepackage{cite,url}

\usepackage{amsmath}
\usepackage{amssymb}

\usepackage{mathrsfs}
\usepackage{amsfonts}

\usepackage{algorithm2e}

\let\savedalgorithm\algorithm
\let\savedendalgorithm\endalgorithm

\usepackage{algorithm}
 \usepackage[margin=1.3in]{geometry}
 \usepackage{setspace}
\usepackage{graphicx}
\usepackage{amsmath}
\usepackage{amssymb}
\usepackage{amsfonts}
\usepackage{mathrsfs,eucal}
\usepackage{bm}

\begin{document}

\markboth{A Survey of Appearance Models in Visual Object Tracking}{X. Li et al.}

\title{A Survey of Appearance Models in Visual Object Tracking}

\author{Xi Li$^{2,1}$, Weiming Hu$^{2}$, Chunhua Shen$^{1}$,
Zhongfei Zhang$^{3}$, Anthony Dick$^{1}$,
Anton van den Hengel$^{1}$
\vspace{.5cm}
\\
$^1\,$The University of Adelaide, Australia\\
$^2\,$Institute of Automation, Chinese Academy of Sciences, China\\
$^3\,$State University of New York, Binghamton, USA
}

\begin{abstract}

    Visual object tracking is a significant computer vision task which
    can be applied to many domains such as visual surveillance, human
    computer interaction, and video compression.  Despite extensive
    research on this topic, it still suffers from difficulties in
    handling complex object appearance changes caused by factors such
    as illumination variation, partial occlusion, shape deformation,
    and camera motion.  Therefore, effective modeling of the 2D
    appearance of tracked objects is a key issue for the success of a
    visual tracker.
    In the literature,  researchers have proposed a variety of 2D
    appearance models.

    To help readers swiftly learn the recent advances in 2D appearance
    models for visual object tracking, we contribute this survey,
    which provides a detailed review of the existing 2D appearance
    models.  In particular, this survey takes a module-based
    architecture  that enables readers to easily grasp the key points
    of visual object tracking.  In this survey, we first decompose the
    problem of appearance modeling into two different processing
    stages: visual representation and statistical modeling.  Then,
    different 2D appearance models are categorized and discussed with
    respect to their composition modules.  Finally, we address several
    issues of interest as well as the remaining challenges for
    future research on this topic.

    The contributions of this survey are four-fold. First, we review
    the literature of visual representations according to their
    feature-construction mechanisms (i.e., local and global).  Second,
    the existing statistical modeling schemes for
    tracking-by-detection are reviewed according to their
    model-construction mechanisms: generative, discriminative, and
    hybrid generative-discriminative.  Third, each type of visual
    representations or statistical modeling techniques is analyzed and
    discussed from a theoretical or practical viewpoint.  Fourth, the
    existing benchmark resources (e.g., source code and video
    datasets) are examined in this survey.

\end{abstract}

\category{I.4.8}{Image Processing and Computer Vision}{Scene Analysis--Tracking}

\terms{Algorithms, Performances}

\keywords{Visual object tracking, appearance model, features, statistical
modeling}

\maketitle

\section{Introduction}

\renewcommand{\vspace}[1]{}

One of the main goals of computer vision is to enable computers to replicate the basic functions of
human vision such as motion perception and scene understanding. To achieve the goal of intelligent
motion perception, much effort has been spent on visual object tracking, which is one of the
most important and challenging research topics in computer vision.
Essentially, the core of visual object tracking
is to robustly estimate the motion state (i.e., location, orientation, size, etc.) of a target
object in each frame of an input image sequence.

In recent years, a large body of research on visual object tracking has been published in the
literature.
Research interest in visual object tracking comes from the fact that it has a wide range of
real-world applications, including visual surveillance, traffic flow monitoring, video compression,
and human-computer interaction.
For example, visual object tracking is successfully applied to monitor human activities
in residential areas, parking lots, and banks (e.g., $\mbox{W}^{4}$ system~\cite{Haritaoglu-Harwood-Davis-PAMI2000}
and VSAM project~\cite{Collins-Lipton-Kanade-TR2000}).
In the field of traffic transportation,
visual object tracking is also widely used to cope with traffic flow
monitoring~\cite{Coifman-Beymer-Mclauchlan-Malik-TRP1998}, traffic accident
detection~\cite{Tai-Tsang-Lin-Song-IVC2004}, pedestrian
counting~\cite{Masoud-Papanikolopoulos-TranVT2001}, and so on.
Moreover, visual object tracking is utilized by the MPEG-4 video compression standard~\cite{Sikora-CSVT1997} to automatically
detect and track moving objects in videos.
As a result, more encoding bytes are assigned to moving objects while
fewer
encoding bytes are for redundant backgrounds.
Visual object tracking
also has several human-computer interaction applications such as
hand gesture recognition \cite{Pavlovie-Sharma-Huang-PAMI1997} and mobile video
conferencing \cite{Paschalakis-Bober-RTI2004}.

\begin{figure*}[t]
\begin{center}
\includegraphics[width=0.83\linewidth]{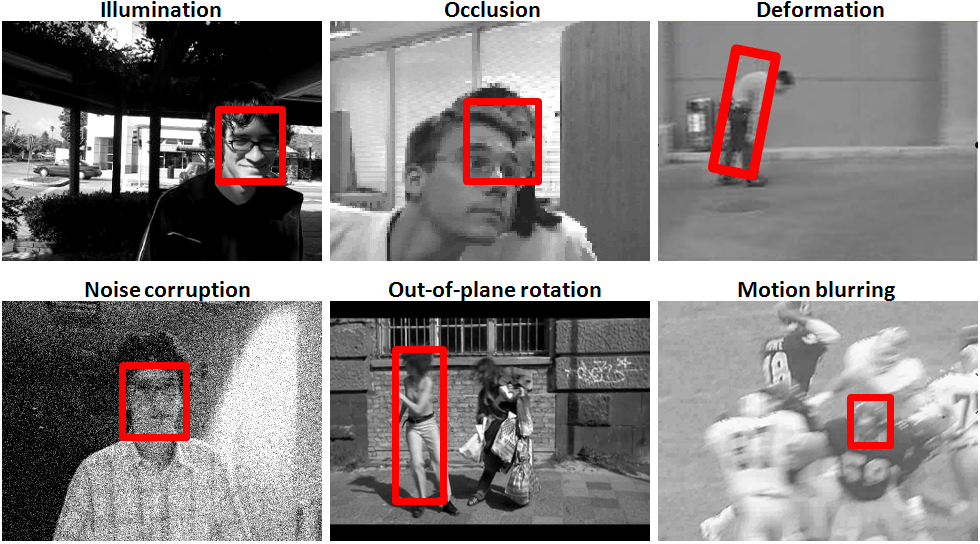}
\end{center}
\vspace{-0.23cm}
   \caption{Illustration of complicated appearance changes in visual object tracking.}
    \label{fig:appearance_change}
\end{figure*}

Note that all the above applications heavily rely  on the
information provided by a
robust visual object tracking method.
If such information is not available, these applications
would be no longer valid.
Therefore, robust visual object tracking
is a key issue to make these applications viable.

\subsection{Overview of visual object tracking}

In general, a typical visual object tracking system is composed of four modules:
object initialization, appearance modeling, motion estimation,
and object localization.

\begin{itemize}
\item {\em Object initialization}. This may be manual or automatic. Manual initialization is performed by users to annotate object locations with
bounding boxes or ellipses. In contrast, automatic initialization is usually achieved by object detectors (e.g., face or human detectors).
\item {\em Appearance modeling}. This generally consists of two components: visual representation and
statistical modeling.  Visual representation focuses on how to construct robust object descriptors
using different types of visual features.  Statistical modeling concentrates on how to build
effective mathematical models for object identification  using statistical learning techniques.
\item {\em  Motion estimation}. This is formulated as a dynamic state estimation
problem:
$x_{t} = f(x_{t-1}, v_{t-1})$ and
$z_{t} = h(x_{t}, w_{t})$,
where $x_{t}$ is the current state,
$f$ is the state evolution function, $v_{t-1}$ is the evolution process noise, $z_{t}$ is the
current observation, $h$ denotes the measurement function, and $w_{t}$ is the measurement noise.
The task of motion estimation is usually completed by utilizing predictors such as linear
regression techniques~\cite{Ellis-IJCV2010}, Kalman filters~\cite{Kalman-Book}, or particle
filters~\cite{Isard-Blake-IJCV1998,Arulampalam-Maskell-Gordon-Clapp-tsp2002,SHEN05ICIP}.
\item {\em  Object localization}. This is performed by a greedy search or maximum a posterior estimation based on motion
estimation.
\end{itemize}

\subsection{Challenges in developing robust appearance models}

\begin{figure*}[t]
\begin{center}
\includegraphics[width=0.8\linewidth]{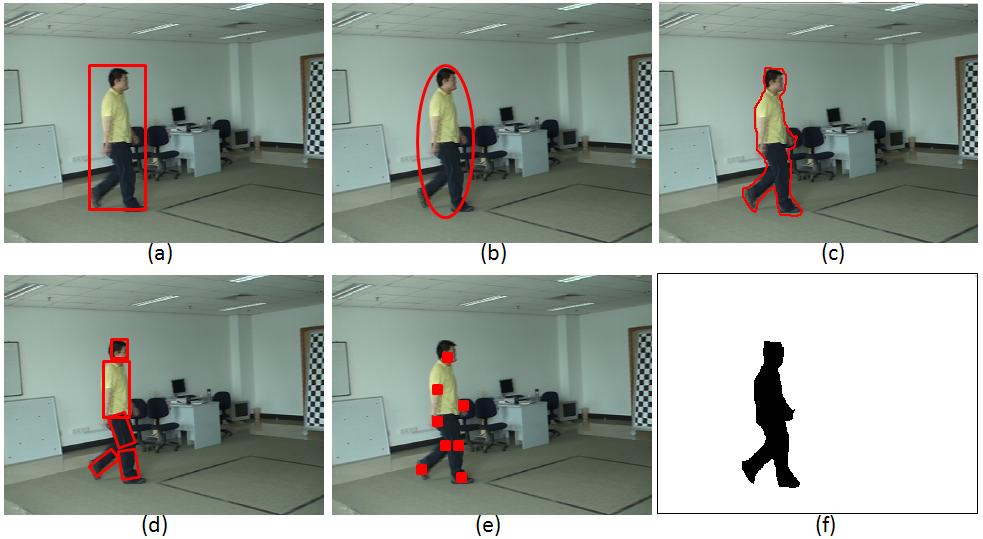}
\end{center}
\vspace{-0.33cm}
   \caption{Illustration of object tracking forms. (a) bounding box,
   (b) ellipse, (c) contour, (d) articulation block, (e) interest point,
   (f) silhouette.}
    \label{fig:what-to-track}
\end{figure*}

Many issues have made robust visual object tracking very challenging,
including (i) low-quality camera sensors (e.g., low frame rate, low resolution,
low bit-depth, and color distortion); (ii) challenging factors
(e.g., non-rigid object tracking, small-size object tracking, tracking a varying number of objects, and complicated pose estimation);
(iii) real-time processing requirements;
(iv) object tracking across cameras with non-overlapping views~\cite{javed2008modeling};
and (v) object appearance variations (as shown in Fig.~\ref{fig:appearance_change}) caused by several complicated factors
(e.g., environmental illumination changes, rapid camera motions, full occlusion, noise disturbance, non-rigid
shape deformation, out-of-plane object rotation, and pose variation).
These challenges may cause tracking degradations and even failures.

In order to deal with these challenges, researchers have proposed a wide range of appearance models
using different visual representations and/or statistical modeling techniques. These appearance models usually
focus on different problems in visual object tracking, and thus have different properties and
characteristics. Typically, they attempt to answer the following questions:
\begin{itemize}
\item What to track (e.g., bounding box, ellipse, contour, articulation block, interest point, and silhouette, as shown in Fig.~\ref{fig:what-to-track})?
\item What visual representations are appropriate and robust for visual object tracking?
\item What are the advantages or disadvantages
of different visual representations for different tracking tasks?
\item Which types of statistical learning
schemes are suitable for visual object tracking?
\item What are the properties or characteristics of these
statistical learning schemes during visual object tracking?
\item How should the camera/object motion be modeled
in the tracking process?
\end{itemize}
The answers to these questions rely heavily on the specific
context/environment of the tracking task and the tracking information available to users.
Consequently, it is necessary to categorize these appearance models into several task-specific
categories and discuss in detail the representative appearance models of each category.
Motivated by this consideration, we provide a survey to help readers acquire valuable tracking knowledge
and choose the most suitable appearance model for their particular tracking tasks. Furthermore, we
examine several interesting issues for developing new appearance models.

\begin{figure*}[t]
\hspace{-0.19cm}
\includegraphics[width=1\linewidth]{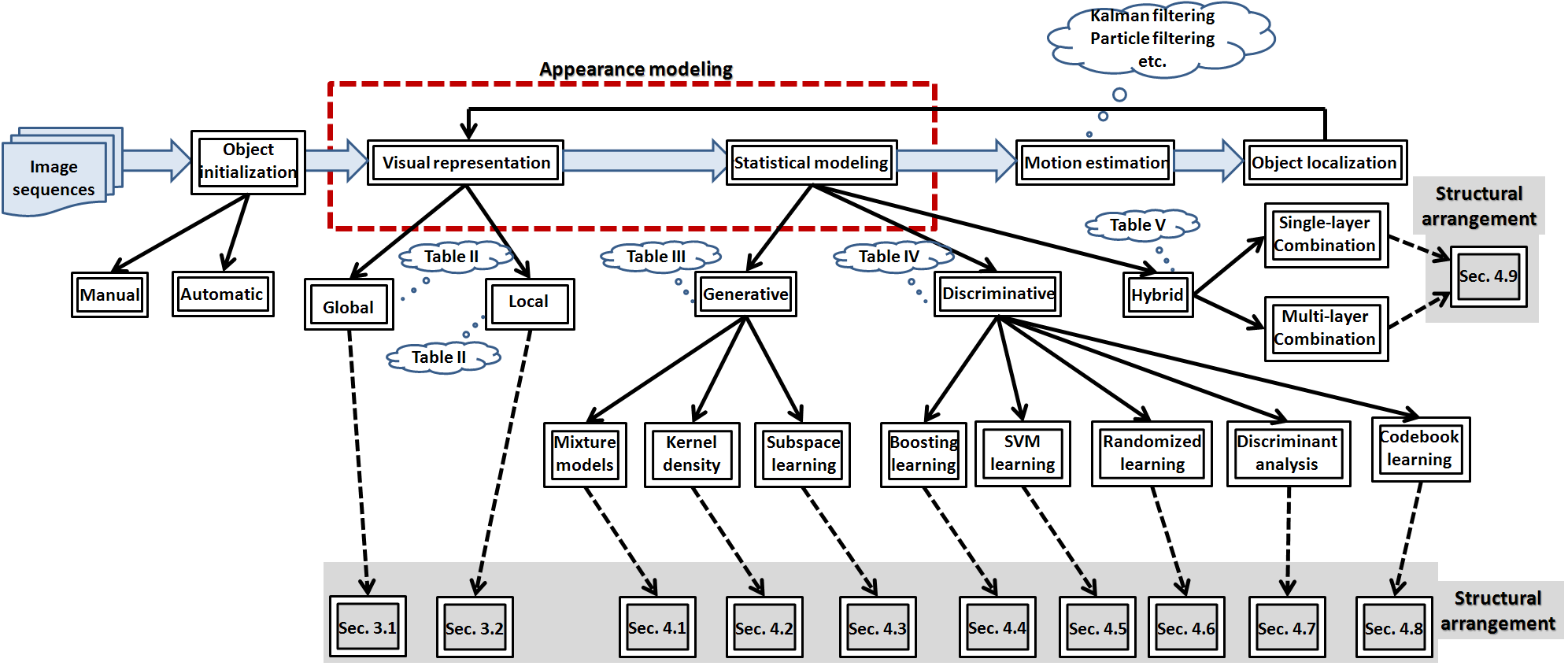}
\vspace{-0.11cm}
   \caption{The organization of this survey.}
    \label{fig:architecture}
\end{figure*}

\section{Organization of this survey}

Fig.~\ref{fig:architecture} shows the
organization of this survey, which is composed of two modules:
visual representation and statistical modeling. The visual representation
module concentrates on how to robustly describe the spatio-temporal
characteristics of object appearance. In this module,
a variety of visual representations are discussed, as illustrated
by the tree-structured taxonomy in the left part of Fig.~\ref{fig:architecture}.
These visual representations can capture various
visual information
at different levels (i.e., local and global).
Typically, the local visual representations encode the local statistical  information  (e.g., interest point) of an image region,
while the global visual representations reflect the  global statistical characteristics (e.g., color histogram)
of an image region.
For a clear illustration of this module, a detailed literature review of visual representations is given in Sec.~\ref{sec:feature_description}.

\begin{table*}[t]
\vspace{-0.3cm}
\caption{{Summary of related literature surveys}}
\label{tab:survey_list}
\footnotesize
\hspace{0.2cm}
\scalebox{0.92}{
\begin{tabular}{c|*{4}{c}}\hline
Authors  & Topic   & Journal/conference title \\\hline\hline
\cite{Geronimo-Lopez-Sappa-Graf-PAMI2010}  & Pedestrian Detection & IEEE Trans. on
PAMI.  \\ \hline
\cite{Candamo-Shreve-Goldgof-Sapper-Kasturi-TransITS2010} & Human Behavior Recognition & IEEE Trans. on
Intelligent Transportation Systems  \\ \hline
\cite{Cannons-TR2008}  & Visual Tracking & Technical Report  \\ \hline
\cite{Zhan-Monekosso-Remagnino-Velastin-Xu-MVA2008} & Crowd analysis & Machine Vision Application \\ \hline
\cite{Kang-Deng-ICCIS2007}   & Intelligent Visual Surveillance & IEEE/ACIS
Int. Conf. Comput. Inf. Sci. \\ \hline
\cite{Yilmaz-Javed-Shah-ACMSurvey2006} & Visual object tracking &ACM Computing Survey  \\ \hline
\cite{Forsyth-Arikan-Ikemoto-Brien-Ramanan2006}  &  Human Motion Analysis & Found. Trends
Comput. Graph. Vis. \\ \hline
\cite{Sun-Bebis-Miller-PAMI2006}  & Vehicle Detection & IEEE Trans. on PAMI.  \\ \hline
\cite{Hu-Tan-Wang-Maybank-TransSMC2004}  & Object Motion and Behaviors & IEEE Trans. on
Syst., Man, Cybern. C, Appl. Rev.  \\ \hline
\cite{Arulampalam-Maskell-Gordon-Clapp-tsp2002}  & Bayesian Tracking & IEEE Trans. on Signal Processing
\\\hline
\end{tabular}
}
\end{table*}

As shown in the right part of Fig.~\ref{fig:architecture}, the statistical modeling module is inspired by the
tracking-by-detection idea, and thus
focuses on using different
types of statistical learning schemes to learn a
robust statistical model for object detection, including generative, discriminative,
and hybrid generative-discriminative ones.  In this module, various tracking-by-detection methods
based on different statistical modeling
techniques
are designed to facilitate different statistical properties of the
object/non-object class.
For a clear illustration of this module, a detailed literature review of
statistical modeling schemes for tracking-by-detection is given
in Sec.~\ref{sec:statistical_modeling}.

Moreover,  a number of source codes and video datasets for visual object tracking are examined to
make them easier for readers to conduct tracking experiments in Sec.~\ref{sec:benchmark}.
Finally, the survey is concluded in Sec.~\ref{sec:Conclusion}. In particular, we additionally
address several interesting issues for the future research
in Sec.~\ref{sec:Conclusion}.

\subsection{Main differences from other related surveys}
\label{sec:related_work}

In the recent literature, several related surveys
(e.g., \cite{Geronimo-Lopez-Sappa-Graf-PAMI2010,Candamo-Shreve-Goldgof-Sapper-Kasturi-TransITS2010,Cannons-TR2008,Zhan-Monekosso-Remagnino-Velastin-Xu-MVA2008,Kang-Deng-ICCIS2007,Yilmaz-Javed-Shah-ACMSurvey2006,Forsyth-Arikan-Ikemoto-Brien-Ramanan2006,Sun-Bebis-Miller-PAMI2006,Hu-Tan-Wang-Maybank-TransSMC2004,Arulampalam-Maskell-Gordon-Clapp-tsp2002}) of
visual object tracking  have been made to
investigate the state-of-the-art tracking algorithms and their potential
applications, as listed in Tab.~\ref{tab:survey_list}. Among these surveys, the topics of the
surveys~\cite{Cannons-TR2008,Yilmaz-Javed-Shah-ACMSurvey2006} are closely related to this
paper.  Specifically, both of the
surveys~\cite{Cannons-TR2008,Yilmaz-Javed-Shah-ACMSurvey2006} focus on low-level tracking
techniques using different visual features or statistical learning techniques, and thereby give very
comprehensive and specific technical contributions.

The main differences between these two surveys~\cite{Cannons-TR2008,Yilmaz-Javed-Shah-ACMSurvey2006} and this survey are summarized as follows.
First, this survey focuses on  the 2D appearance modeling for visual object tracking.  In
comparison, the surveys of \cite{Cannons-TR2008,Yilmaz-Javed-Shah-ACMSurvey2006} concern all the
modules shown in Fig.~\ref{fig:architecture}.  Hence, this survey is more intensive while
the surveys of \cite{Cannons-TR2008,Yilmaz-Javed-Shah-ACMSurvey2006} are more comprehensive.
Second, this survey provides a more detailed analysis of
various appearance models.
Third, the survey of \cite{Yilmaz-Javed-Shah-ACMSurvey2006} splits visual object tracking into
three categories: point tracking, kernel tracking, and silhouette tracking (see Fig. 7
in~\cite{Yilmaz-Javed-Shah-ACMSurvey2006} for details); the survey of \cite{Cannons-TR2008} gives a very detailed and comprehensive review
of each tracking issue in visual object tracking. In contrast to these
two surveys,
this survey is formulated as a general module-based architecture
(shown in Fig.~\ref{fig:architecture}) that enables readers to easily grasp the key points of
visual object tracking.
Fourth, this survey investigates a large number of state-of-the-art appearance models
which make use of novel visual features and statistical learning techniques.
In comparison, the surveys~\cite{Cannons-TR2008,Yilmaz-Javed-Shah-ACMSurvey2006}
pay more attention to classic and fundamental appearance models used
for visual object tracking.

\begin{table*}[t]
\vspace{-0.5cm}
\caption{{Summary of representative visual representations}}
\label{tab:feature_descriptor}
\hspace{-0.0cm}
\scalebox{0.66}{
\begin{tabular}{c|c|c|c}\hline
Item No.&
\begin{tabular}{c}
References  \\
\end{tabular}
 &
\begin{tabular}{c}
Global/local \\
\end{tabular}
&
\begin{tabular}{c}
Visual representations
\end{tabular}
 \\\hline\hline
1&
\begin{tabular}{c}
\cite{Ho-Lee-Yang,Li-Xu,Limy-Ross17}\\
\end{tabular}
 & Global &
\begin{tabular}{c}
Vector-based raw pixel representation
\end{tabular}
\\ [0 ex] \hline

2 &
\begin{tabular}{c}
\cite{lixi-iccv2007}\\
\end{tabular}
 & Global &
\begin{tabular}{c}
Matrix-based raw pixel representation
\end{tabular}
\\ \hline

3&
\begin{tabular}{c}
\cite{Wang-Suter-Schindler-PAMI2007}\\
\end{tabular}
& Global &
\begin{tabular}{c}
Multi-cue raw pixel representation\\
(i.e., color, position, edge)
\end{tabular}
\\ \hline

4&
\begin{tabular}{c}
\cite{Werlberger-Trobin-Pock-Wedel-BMVC2009,Santner-Leistner-Saffari-Pock-Bischof-cvpr2010}\\
\end{tabular}
& Global &
\begin{tabular}{c}
Optical flow representation\\
(constant-brightness-constraint)
\end{tabular}
\\ \hline

5&
\begin{tabular}{c}
\cite{Black-Anandan-CVIU1996,Wu-Fan-CVPR2009}\\
\end{tabular}
& Global &
\begin{tabular}{c}
Optical flow representation\\
(non-brightness-constraint)
\end{tabular}
\\ \hline

6&
\begin{tabular}{c}
\cite{Bradski-WACV1998}\\
\cite{Comaniciu-Ramesh-Meer-TPAMI,Zhao-Brennan-Tao-PAMI2010}\\
\end{tabular}
 & Global &
\begin{tabular}{c}
Color histogram representation \\
\end{tabular}
\\ \hline

7&
\begin{tabular}{c}
\cite{Georgescu-Meer-PAMI2004}\\
\end{tabular}
 & Global &
\begin{tabular}{c}
Multi-cue spatial-color histogram representation \\
(i.e., joint histogram in (x, y, R, G, B))
\end{tabular}
\\ \hline

8&
\begin{tabular}{c}
\cite{Adam-Rivlin-Shimshoni-cvpr2006}
\end{tabular}
& Global &
\begin{tabular}{c}
Multi-cue spatial-color histogram representation \\
(i.e., patch-division histogram)
\end{tabular}
\\ \hline

9&
\begin{tabular}{c}
\cite{haralick1973textural,gelzinis2007increasing}
\end{tabular}
& Global &
\begin{tabular}{c}
Multi-cue spatial-texture histogram representation \\
(i.e., Gray-Level Co-occurrence Matrix)
\end{tabular}
\\ \hline

10&
\begin{tabular}{c}
\cite{Haritaoglu-Flickner-CVPR2001}\\
\cite{Ning-Zhang-Zhang-Wu-IJPRAI2009}\\
\end{tabular}
 & Global &
\begin{tabular}{c}
Multi-cue shape-texture histogram representation\\
(i.e., color, gradient, texture)
\end{tabular}
\\ \hline

11&
\begin{tabular}{c}
\cite{Porikli-Tuzel-Meer-cvpr2006,Wu-Wu-Liu-Lu-ICPR2008}\\
\end{tabular}
 & Global &
\begin{tabular}{c}
Affine-invariant\\
covariance representation
\end{tabular}
\\ \hline

12&
\begin{tabular}{c}
\cite{lixi-cvpr2008,hong2010sigma}\\
\cite{wu2012realtip,hu2012single}
\end{tabular}
 & Global &
\begin{tabular}{c}
Log-Euclidean\\
covariance representation
\end{tabular}
\\ \hline

13&
\begin{tabular}{c}
\cite{he2002object,Li-Zhang-Huang-Tan-ICIP2009}
\end{tabular}
 & Global &
\begin{tabular}{c}
Wavelet filtering-based representation
\end{tabular}
\\ \hline

14&
\begin{tabular}{c}
\cite{Paragios-Deriche-PAMI2000,Cremers-PAMI2006}\\
\cite{Allili-Ziou-CVPR2007,sun2011novel}
\end{tabular}
 & Global &
\begin{tabular}{c}
Active contour representation
\end{tabular}
\\ \hline

15&
\begin{tabular}{c}
\cite{lin2007hierarchical}
\end{tabular}
 & Local &
\begin{tabular}{c}
Local feature-based represnetation\\
(local templates)
\end{tabular}
\\ \hline

16&
\begin{tabular}{c}
\cite{Tang-Tao-TCSVT2008,Zhou-Yuan-Shi-CVIU2009}\\
\end{tabular}
& Local &
\begin{tabular}{c}
Local feature-based represnetation\\
(SIFT features)
\end{tabular}
\\ \hline

17&
\begin{tabular}{c}
\cite{Donoser-Bischof-CVPR2006,Tran-Davis-iccv2007}\\
\end{tabular}
 & Local &
\begin{tabular}{c}
Local feature-based represnetation\\
(MSER features)
\end{tabular}
\\ \hline

18&
\begin{tabular}{c}
\cite{He-Yamashita-Lu-Lao-ICCV2009}
\end{tabular}
 & Local &
\begin{tabular}{c}
Local feature-based represnetation\\
(SURF features)
\end{tabular}
\\ \hline

19&
\begin{tabular}{c}
\cite{Grabner-Grabner-Bischof-cvpr2007,Kim-CVPR2008}\\
\end{tabular}
 & Local &
\begin{tabular}{c}
Local feature-based represnetation\\
(Corner features)
\end{tabular}
\\ \hline

20&
\begin{tabular}{c}
\cite{Collins-Liu-Leordeanu-PAMI2005,Grabner-Bischof-CVPR2006}\\
\cite{Yu-Dinh-Medioni-ECCV2008}\\
\end{tabular}
 & Local &
\begin{tabular}{c}
Local feature-based represnetation\\
(feature pools
of Harr, HOG, LBP etc.)
\end{tabular}
\\ \hline

21&
\begin{tabular}{c}
\cite{Toyama-Hager-IJCV1996}\\
\cite{Mahadevan-Vasconcelos-CVPR2009}\\
\cite{Yang-Yuan-Wu-cvpr2007,Fan-Wu-Dai-ECCV2010}\\
\end{tabular}
& Local &
\begin{tabular}{c}
Local feature-based representations\\
(Saliency detection-based features)
\end{tabular}
\\ \hline

22&
\cite{ren2007tracking,Superpixeltrackingiccv2011}
& Local &
\begin{tabular}{c}
Local feature-based represnetation\\
(Segmentation-based features)
\end{tabular}
\\\hline

\end{tabular}
}
\end{table*}

\subsection{Contributions of this survey} \label{sec:motivation}

The contributions of this survey are as follows.
First, we  review the literature of
visual representations from a feature-construction viewpoint.
Specifically, we hierarchically categorize visual representations into local and global features.
Second, we take a tracking-by-detection criterion for reviewing
the existing statistical modeling schemes.
According to the model-construction mechanisms, these statistical modeling
schemes are roughly classified into three categories: generative, discriminative, and hybrid generative-discriminative.
For each category, different types of statistical learning
techniques for object detection are reviewed and discussed.
Third, we provide a detailed discussion on each type of visual representations or statistical learning techniques with their properties.
Finally, we examine the existing benchmark resources for visual object tracking, including
source codes and databases.

\vspace{-0.0cm}
\section{Visual representation}
\label{sec:feature_description}
\vspace{-0.05cm}
\subsection{Global visual representation}

A global visual representation reflects the  global statistical characteristics
of object appearance. Typically, it can be investigated in the following
main aspects: (i) raw pixel representation; (ii) optical flow representation;
(iii) histogram representation; (iv)covariance representation;
(v) wavelet filtering-based representation; and (vi)active contour representation.
Tab.~\ref{tab:feature_descriptor} lists several representative tracking
methods using global visual representations (i.e., Rows 1-14).

\begin{itemize}
\item Raw pixel representation. As the most fundamental features in computer vision,
raw pixel values
are widely used in visual object tracking because of their simplicity and efficiency.
Raw pixel representation directly utilizes the raw
color or intensity values of the image pixels to represent the object regions.
Such a representation is simple and efficient for fast object tracking.
In the literature, raw pixel representations are usually constructed
in the following two forms: vector-based~\cite{Silveira-Malis-cvpr2007,Ho-Lee-Yang,Li-Xu,Limy-Ross17}
and matrix-based~\cite{lixi-iccv2007,Wen-Gao-Li-Tao-SMC2009,Hu-Li-IJCV2010,Wang-Gu-Shi-ICASSP2007,Li-Liang-Huang-Jiang-Gao-ICIP2008}.
The vector-based representation directly flattens an image region into a high-dimensional vector, and
often suffers from a small-sample-size problem.
Motivated by attempting to alleviate the small-sample-size problem, the matrix-based representation directly utilizes 2D matrices
or higher-order tensors as the basic data units for object description due to its relatively low-dimensional property.

\hspace{0.18cm} However, raw pixel information alone is not enough for robust visual object tracking.
Researchers attempt to embed  other visual cues (e.g., shape or texture) into the raw pixel representation.
Typically, the color features are enriched
by fusing other visual information such as edge~\cite{Wang-Suter-Schindler-PAMI2007} and
texture~\cite{Allili-Ziou-CVPR2007}.

\item Optical flow representation. In principle, optical flow represents a dense field of
displacement vectors of each pixel inside an image region,
and is commonly used to capture the spatio-temporal
motion information of an object.
Typically, optical flow
has two branches:
constant-brightness-constraint (CBC) optical
flow~\cite{Lucas-Kanade-IJCAI1981,Horn-Schunck-AI1981,Werlberger-Trobin-Pock-Wedel-BMVC2009,Sethi-Jain-PAMI1987,Salari-Sethi-PAMI1990,Santner-Leistner-Saffari-Pock-Bischof-cvpr2010}
and non-brightness-constraint (NBC) optical flow~\cite{Black-Anandan-CVIU1996,Sawhney-Ayer-PAMI1996,Hager-Belhumeur-PAMI1998,Bergen-Burt-Hingorani-Peleg-PAMI1992,Irani-ICCV1999,Wu-Fan-CVPR2009}.
The CBC optical flow has a constraint on brightness constancy while the NBC optical flow
deals with the situations with varying lighting conditions.

\item Histogram representation. Histogram representations are popular in visual object tracking because of their effectiveness and
efficiency in capturing the distribution characteristics of visual features inside the object
regions.  In general, they have two branches:  single-cue and multi-cue.

\hspace{0.18cm} (i)  A single-cue histogram representation often constructs
a histogram to capture the distribution information inside an object region.
For example, Bradski~\citeyear{Bradski-WACV1998} uses a color histogram in the Hue Saturation Value (HSV) color space for object
representation, and then embeds the color histogram into a continuously adaptive mean shift (CAMSHIFT)
framework for object tracking.
However, the direct use of color histogram may result in the loss of spatial
information.  Following the work in~\cite{Bradski-WACV1998}, Comaniciu et
al.~\citeyear{Comaniciu-Ramesh-Meer-TPAMI} utilize
a spatially weighted color histogram in the RGB color space for visual representation,
and subsequently embed the spatially weighted color histogram
into a mean shift-based tracking framework for object state inference.
Zhao et al.~\citeyear{Zhao-Brennan-Tao-PAMI2010}
convert the problem of object tracking into that of
matching the RGB color distributions across frames.
As a result, the task of object localization
is taken by
using a fast differential EMD (Earth Mover's Distance)
to compute the similarity between the
color distribution of the learned target and the color distribution of a candidate region.

\hspace{0.18cm} (ii) A multi-cue histogram representation aims to encode more information
to enhance the robustness of visual representation. Typically, it
contains three main components:
a) spatial-color; b) spatial-texture; c) shape-texture;

\hspace{0.18cm} For a), two strategies are adopted, including joint spatial-color modeling and
patch-division. The goal of joint spatial-color modeling is to describe the
distribution properties of object appearance
in a joint spatial-color space (e.g., (x, y, R, G, B) in~\cite{Yang-Duraiswami-Davis-CVPR2005,Georgescu-Meer-PAMI2004,Birchfield-Rangarajan-CVPR2005}).
The patch-division strategy is to  encode the spatial information into the appearance models by splitting the tracking
region into a set of patches~\cite{Adam-Rivlin-Shimshoni-cvpr2006,Nejhum-Ho-Yang-CVIU2010}. By considering the geometric relationship between patches, it is capable of capturing the spatial
layout information.  For example, Adam et
al.~\cite{Adam-Rivlin-Shimshoni-cvpr2006} construct a patch-division
visual representation with a histogram-based feature description for object tracking, as shown in Fig.~\ref{fig:patch_division}.  The final tracking position is
determined by combining the vote maps of all patches (represented by grayscale histograms). The
combination mechanism can eliminate the influence of the outlier vote maps caused by occlusion.
For the computational efficiency, Porikli~\citeyear{porikli2005integral}
introduces a novel concept of an integral histogram
to compute the histograms of all possible target
regions in a Cartesian data space. This greatly accelerates
the speed of histogram matching in the process of mean shift tracking.

\begin{figure*}[t]
\begin{center}
\includegraphics[width=0.58\linewidth]{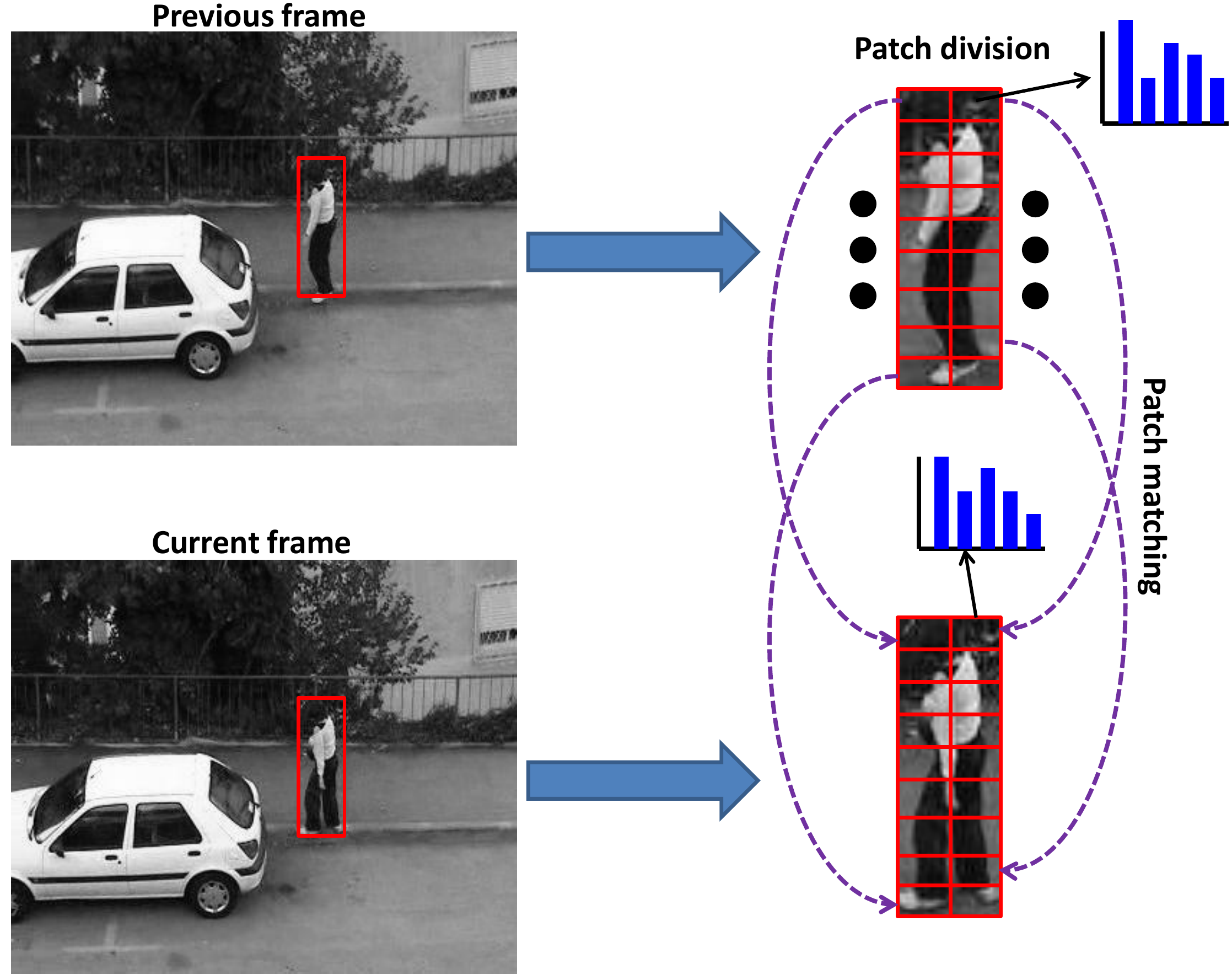}
\end{center}
\vspace{-0.23cm}
\caption{Illustration of patch-division visual representation (from [Adam et al. 2006], $\copyright$2006 IEEE).
The left part shows the previous and current frames, and the right part displays the patch-wise histogram matching
process between two image regions.} \vspace{-0.1cm}
\label{fig:patch_division}
\end{figure*}

\hspace{0.18cm} For b), an estimate of the joint spatial-texture probability is made
to capture the distribution information on object appearance. For example, Haralick et al.~\citeyear{haralick1973textural}
propose a spatial-texture histogram representation called Gray-Level Co-occurrence Matrix (GLCM),
which encodes the co-occurrence information on pairwise intensities
in a specified direction and distance. Note that the GLCM in~\cite{haralick1973textural}
needs to tune different distance parameter values before selecting
the best distance parameter value by experimental evaluations.
Following the work in~\cite{haralick1973textural},
Gelzinis et al.~\cite{gelzinis2007increasing} propose a GLCM-based histogram
representation that does not need to
carefully select an appropriate distance parameter value.
The proposed histogram
representation gathers the information on the co-occurrence matrices computed for several distance parameter values.

\hspace{0.18cm} For c), the shape or texture information on object appearance is incorporated into the histogram
representation for robust visual object tracking. For instance,
Haritaoglu and Flickner~\citeyear{Haritaoglu-Flickner-CVPR2001} incorporate the gradient or edge information
into the color histogram-based visual representation.  Similar
to~\cite{Haritaoglu-Flickner-CVPR2001}, Wang and Yagi~\citeyear{Wang-Yagi-TIP2008} construct an visual representation using color and shape cues.
The color cues are composed of color histograms in three different color spaces: RGB, HSV, and
normalized $rg$. The shape cue is described by gradient orientation histograms.
To exploit the textural information of the object, Ning et al.~\citeyear{Ning-Zhang-Zhang-Wu-IJPRAI2009}
propose a joint color-texture histogram for visual representation.
The local binary pattern (LBP) technique is employed to identify the key points in the object regions.
Using the identified key points, they build a confidence mask
for joint color-texture feature selection.

\item Covariance representation. In order to capture the correlation information of object appearance, covariance matrix
representations are proposed for visual representation
in~\cite{Porikli-Tuzel-Meer-cvpr2006,Tuzel-Porikli-Meer-ECCV2006}.  According to the
Riemannian metrics mentioned in~\cite{lixi-cvpr2008,hu2012single}, the covariance matrix representations can
be divided into two branches: affine-invariant Riemannian metric-based and Log-Euclidean Riemannian
metric-based.

\hspace{0.18cm} (i) The Affine-invariant Riemannian metric~\cite{Porikli-Tuzel-Meer-cvpr2006,Tuzel-Porikli-Meer-ECCV2006}
is based on the following distance measure:
$
\rho(\mathbf{C}_{1}, \textbf{C}_{2}) =
        \sqrt{\sum_{j=1}^{d}\mbox{ln}^{2}\lambda_{j}(\mathbf{C}_{1}, \mathbf{C}_{2})},
$
where $\{\lambda_{j}(\mathbf{C}_{1}, \mathbf{C}_{2})\}_{j=1}^{d}$ are the generalized
eigenvalues of the two covariance matrices $\mathbf{C}_{1}$ and  $\mathbf{C}_{2}$:
$\lambda_{j}\mathbf{C}_{1}\mathbf{x}_{j} = \mathbf{C}_{2}\mathbf{x}_{j}$, $j\in\{1,\ldots,d\}$, and $\mathbf{x}_{j}$ is the $j$-th generalized
eigenvector. Following the work in~\cite{Porikli-Tuzel-Meer-cvpr2006,Tuzel-Porikli-Meer-ECCV2006},
Austvoll and Kwolek~\citeyear{Austvoll-Kwolek-CVG2010}
use the covariance matrix inside a region to detect whether the feature occlusion events take place.
The detection task can be completed by comparing the covariance matrix-based distance measures
in a particular window around the occluded key point.

\begin{figure*}[t]
\begin{center}
\includegraphics[width=0.6\linewidth]{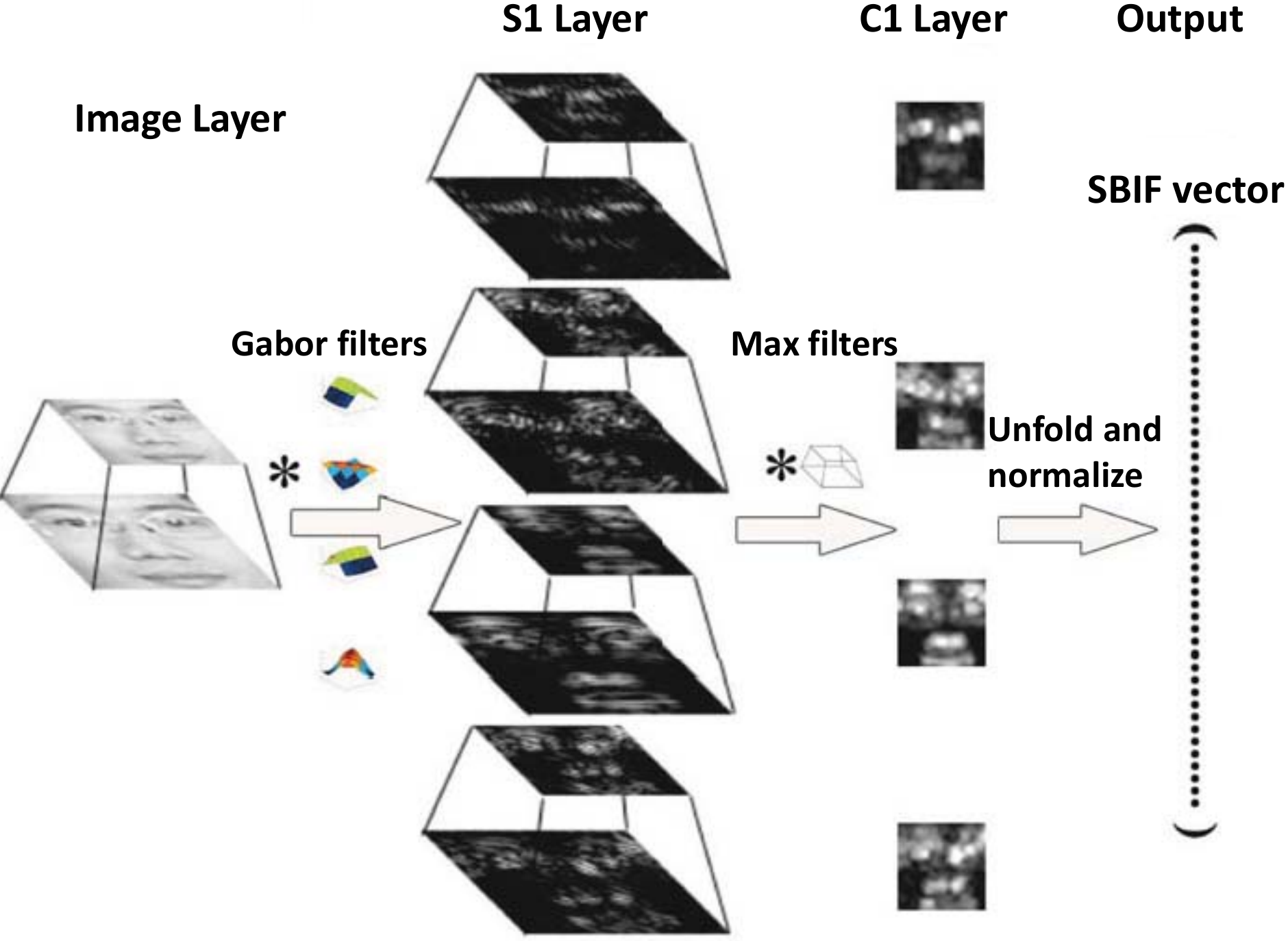}
\end{center}
\vspace{-0.3cm}
   \caption{Illustration of three-layer Gabor features (from [Li et al. 2009], $\copyright$2009 IEEE). The first column shows
   the grayscale face images aligned in a spatial pyramid way (i.e., image layer); the second column the
   Gabor energy maps (containing rich orientation
   and spatial frequency information in the image pyramid) obtained by Gabor filtering (i.e., S1 layer); the third
   column exhibits
   the response of applying max filtering (returning
   the maximum value; and tolerant to local
   distortions) to the Gabor energy maps; and the last column
   plots the final feature vector after unfolding and normalization.
   }
    \label{fig:gabor-representation} \vspace{-0.1cm}
\end{figure*}

\hspace{0.18cm} (ii) The Log-Euclidean Riemannian metric~\cite{Arsigny-Fillard-Pennec-Ayache-SIAM2006} formulates
the distance measure between two covariance matrices in a Euclidean
vector space. Mathematically, the Log-Euclidean Riemannian metric for two covariance matrices $\textbf{C}_{i}$
and $\textbf{C}_{j}$ is formulated as:
$\mbox{d}(\textbf{C}_{i}, \textbf{C}_{j}) = \|\log(\textbf{C}_{i})-\log(\textbf{C}_{j})\|$
where $\log$ is the matrix logarithm operator.
For the descriptive convenience, the covariance matrices under the Log-Euclidean
Riemannian metric are referred to as the Log-Euclidean covariance matrices.
Inspired by~\cite{Arsigny-Fillard-Pennec-Ayache-SIAM2006}, Li et al.~\citeyear{lixi-cvpr2008}
employ the Log-Euclidean covariance matrices of image features for visual representation.
Since the Log-Euclidean covariance matrices lie in a Euclidean
vector space, their mean can be easily computed as the standard arithmetic mean.
Due to this linear property, classic subspace learning techniques (e.g., principal component analysis) can
be directly applied onto the Log-Euclidean covariance matrices.
Following the work in~\cite{lixi-cvpr2008,hu2012single}, Wu et al.~\citeyear{Wu-Cheng-Wang-Lu-iccv2009,wu2012realtip}
extend the tracking problem of using 2D Log-Euclidean covariance matrices to
that of using higher-order tensors, and aim to
incrementally learn a low-dimensional covariance
tensor representation.
Inspired by~\cite{lixi-cvpr2008,hu2012single}, Hong et al.~\citeyear{hong2010sigma} propose
a simplified covariance region descriptor (called Sigma set), which
comprises the lower triangular matrix
square root (obtained by Cholesky factorization) of the covariance matrix (used in~\cite{lixi-cvpr2008}).
The proposed covariance region descriptor characterizes the second order statistics of object
appearance by a set of vectors. Meanwhile, it retains the advantages of
the  region covariance descriptor~\cite{Porikli-Tuzel-Meer-cvpr2006}, such as
low-dimensionality, robustness to noise and illumination variations, and good discriminative power.

\item Wavelet filtering-based representation. In principle, a wavelet filtering-based representation
takes advantage of wavelet transforms to filter the object region
in different scales or directions. For instance, He et al.~\citeyear{he2002object}
utilize a
2D
Gabor wavelet transform (GWT) for visual representation.
Specifically, an object is represented by several feature points
with high GWT coefficients.
Moreover, Li \emph{et al.}~\citeyear{Li-Zhang-Huang-Tan-ICIP2009} propose a tracking algorithm based on
three-layer simplified biologically inspired (SBI) features (i.e., image layer, S1 layer, and C1
layer).  Through the flattening operations on the
four Gabor energy maps in the C1 layer, a unified SBI feature vector is returned to encode the rich
spatial frequency information, as  shown in Fig.~\ref{fig:gabor-representation}.

\item Active contour representation. In order to track the nonrigid objects, active contour representations
have been widely used in recent years~\cite{Paragios-Deriche-PAMI2000,Cremers-PAMI2006,Allili-Ziou-CVPR2007,Vaswani-Rathi-Yezzi-Tannenbaum-TIP2008,sun2011novel}.
Typically, an active contour representation (shown in Fig.~\ref{fig:contour-representation}) is defined as a signed
distance map $\mathbf{\Phi}$: \vspace{-0.15cm}
\begin{equation}
\mathbf{\Phi}(x, y)=
\left\{
\begin{array}{cc}
0 & (x, y)\in C\\
d(x, y, C) & (x, y)\in R_{out}\\
-d(x, y, C) & (x, y)\in R_{in}\\
\end{array}
\right. \vspace{-0.15cm}
\end{equation}
where $R_{in}$ and $R_{out}$ respectively denote the regions inside
and outside the contour C,  and $d(x, y, C)$ is a function returning the smallest Euclidean
distance from point $(x, y)$ to the contour $C$.
Moreover, an active contour representation is associated with a energy function which comprises
three terms: internal energy,
external energy, and shape energy.
The internal energy term reflects the internal constraints on the object contour (e.g., the
curvature-based evolution force), the external energy term measures the likelihood of the image data
belonging to the foreground object class, and the shape energy characterizes the shape prior
constraints on the object contour.

\end{itemize}

\begin{figure*}[t]
\begin{center}
\includegraphics[width=0.7\linewidth]{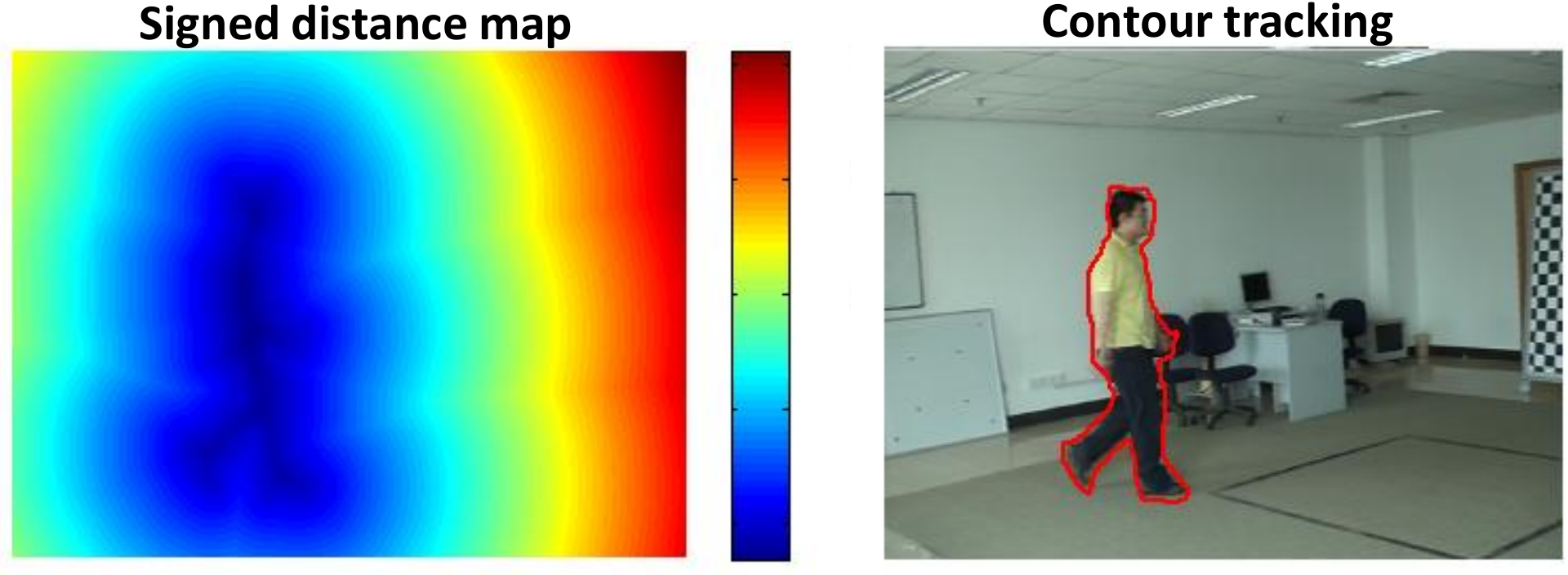}
\end{center}
\vspace{-0.28cm}
   \caption{Illustration of an active contour representation. The left part shows
   the signed distance map of a human contour; and the right part displays the
   contour tracking result.}
    \label{fig:contour-representation} \vspace{-0.23cm}
\end{figure*}

\subsubsection{Discussion}

Without feature extraction, the raw pixel representation is simple and efficient
for visual object tracking. Since only considering the color information on
object appearance,  the raw pixel representation is susceptible to
complicated appearance changes caused by illumination variation.

The constant-brightness-constraint (CBC) optical flow
captures the field information on the translational vectors of
each pixel in a region with the potential
assumption of locally unchanged brightness.
However, the CBC assumption is often invalid in the complicated situations caused by
image noise, illumination fluctuation, and
local deformation.
To address this issue, the non-brightness-constraint optical flow
is developed to introduce more geometric constraints
on the contextual relationship of pixels.

The single-cue histogram representation is capable of efficiently encoding the statistical
distribution information of visual features within the object regions.  Due to its weakness in
characterizing the spatial structural information of tracked objects, it is often affected by
background distractions with similar colors to the tracked objects.  In order to capture more
spatial information, the spatial-color histogram representation is introduced for visual object tracking.
Usually, it encodes the spatial information by either modeling object appearance in a joint
spatial-color feature space or taking a patch-division strategy.  However, the above histogram
representations do not consider the shape or texture information of object appearance.  As a
consequence, it is difficult to distinguish the object from the background with similar color
distributions.  To alleviate this issue, the shape-texture histogram representation is proposed to
integrate shape or texture information (e.g., gradient or edge) into the histogram representation,
leading to the robustness of object appearance variations in illumination and pose.

The advantages of using the covariance matrix representation are as follows:
(i) it can capture the intrinsic self-correlation
properties of object appearance; (ii) it provides an effective way of fusing different
image features from different modalities;
(iii) it is  low-dimensional, leading to the computational efficiency;
(iv) it allows for comparing regions of different sizes or shapes;
(v) it is easy to implement;
(vi) it is robust to illumination changes, occlusion, and shape deformations.
The disadvantages of using the covariance matrix representation are as follows:
(i) it is sensitive to noisy corruption because of taking pixel-wise statistics;
(ii) it loses much useful information such as texture, shape, and location.

A wavelet filtering-based representation
is to encode the local texture information of object appearance
by wavelet transform, which is a convolution with various
wavelet filters. As a result, the wavelet filtering-based representation
is capable of characterizing the statistical properties
of object appearance in multiple scales and directions (e.g., Gabor filtering).

An active contour representation  is designed to cope with
the problem of nonrigid object tracking. Usually,
the active contour representation adopts
the signed distance map to implicitly encode
the boundary information of an object.
On the basis of
level set evolution, the active contour representation can
precisely segment the object with a complicated shape.

\begin{figure*}[t]
\begin{center}
\includegraphics[width=0.8\linewidth]{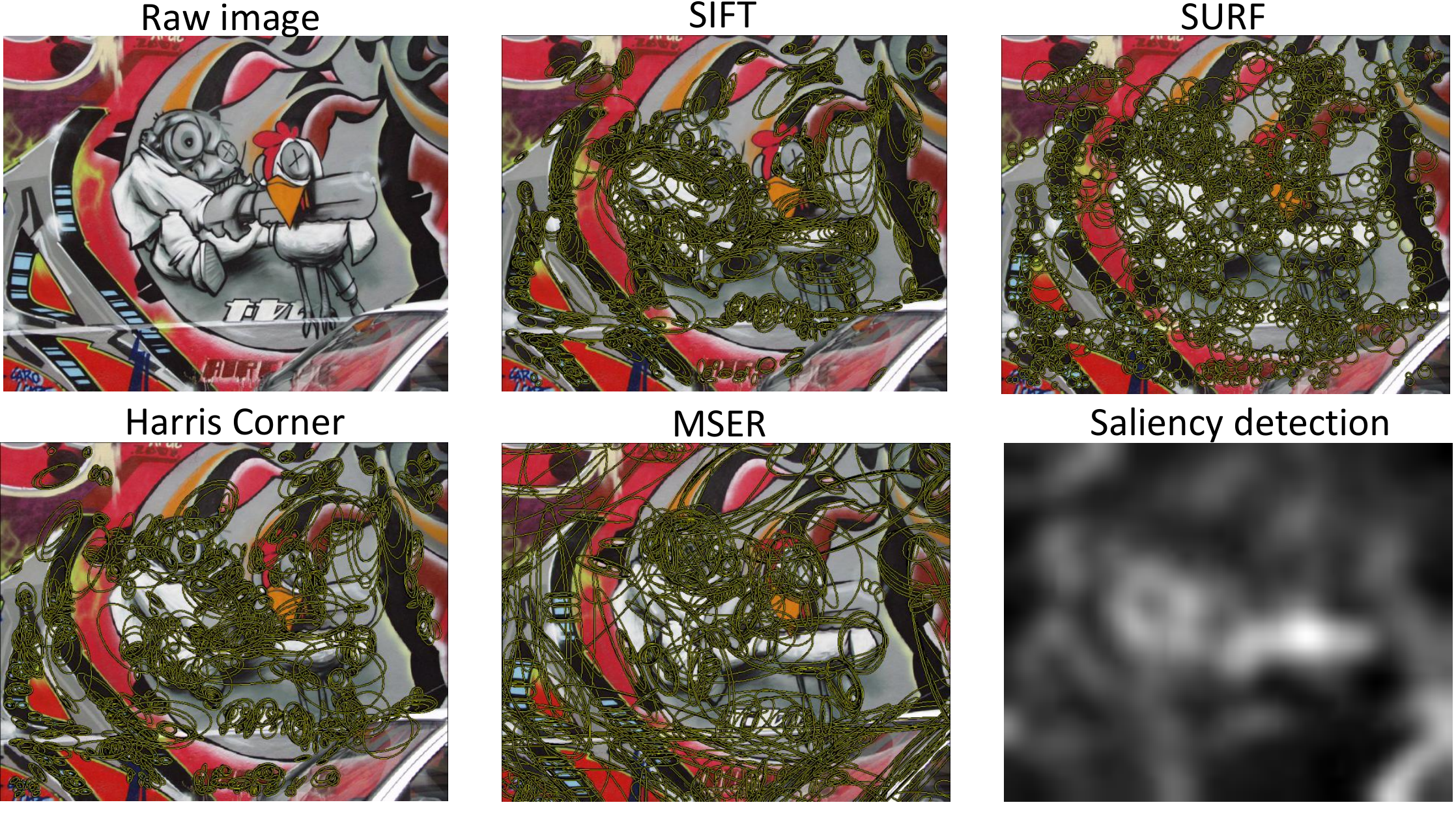}
\end{center}
\vspace{-0.28cm}
   \caption{Illustration of several local features  (extracted by using the software which can be downloaded at 
   \url{http://www.robots.ox.ac.uk/~vgg/research/affine/}
   and \url{http://www.klab.caltech.edu/~harel/share/gbvs.php}).}
    \label{fig:interest_point}
\end{figure*}

\subsection{Local feature-based visual representation}

As shown in Fig.~\ref{fig:interest_point}, local feature-based visual representations mainly utilize
interest points or saliency detection to encode
the object appearance information.
In general, the local features based on the interest points
can be mainly categorized into seven classes: local template-based, segmentation-based,
SIFT-based, MSER-based, SURF-based, corner feature-based, feature pool-based,
and saliency detection-based.
Several representative
tracking methods using local feature-based visual representations
are listed in Rows 15-22 of Tab.~\ref{tab:feature_descriptor}.

\begin{itemize}

\item Local template-based. In general, local template-based visual representations
are to represent an object region using a set of part templates.
In contrast to the global template-based visual representation,
they are able to cope with partial occlusions effectively
and model shape articulations
flexibly. For instance,
a hierarchical part-template shape model is proposed for human detection
and segmentation~\cite{lin2007hierarchical}. The shape model
is associated with a part-template tree that
decomposes
a human body into a set of part-templates.
By hierarchically matching the part-templates with a test image,
the proposed part-template shape model can generate
a reliable set of detection
hypotheses, which are then
put into a global optimization framework for the final human
localization.

\item Segmentation-based. Typically, a segmentation-based visual representation
incorporates the image segmentation cues (e.g., object boundary~\cite{ren2007tracking}) into the process of
object tracking, which leads to reliable tracking results.
Another alternative is based on
superpixel segmentation, which aims to
group pixels into perceptually meaningful atomic regions.
For example, Wang et al.~\citeyear{Superpixeltrackingiccv2011}
construct a local template-based visual representation
with the superpixel segmentation, as shown in Fig.~\ref{fig:superpixel}.  Specifically, the surrounding region
of an object is segmented into several superpixels, each of which
corresponds to a local template.
By building a local template dictionary based on the mean shift clustering,
an object state is predicted by associating
the superpixels of a candidate sample with
the local templates in the dictionary.

\begin{figure*}[t]
\begin{center}
\includegraphics[width=0.8\linewidth]{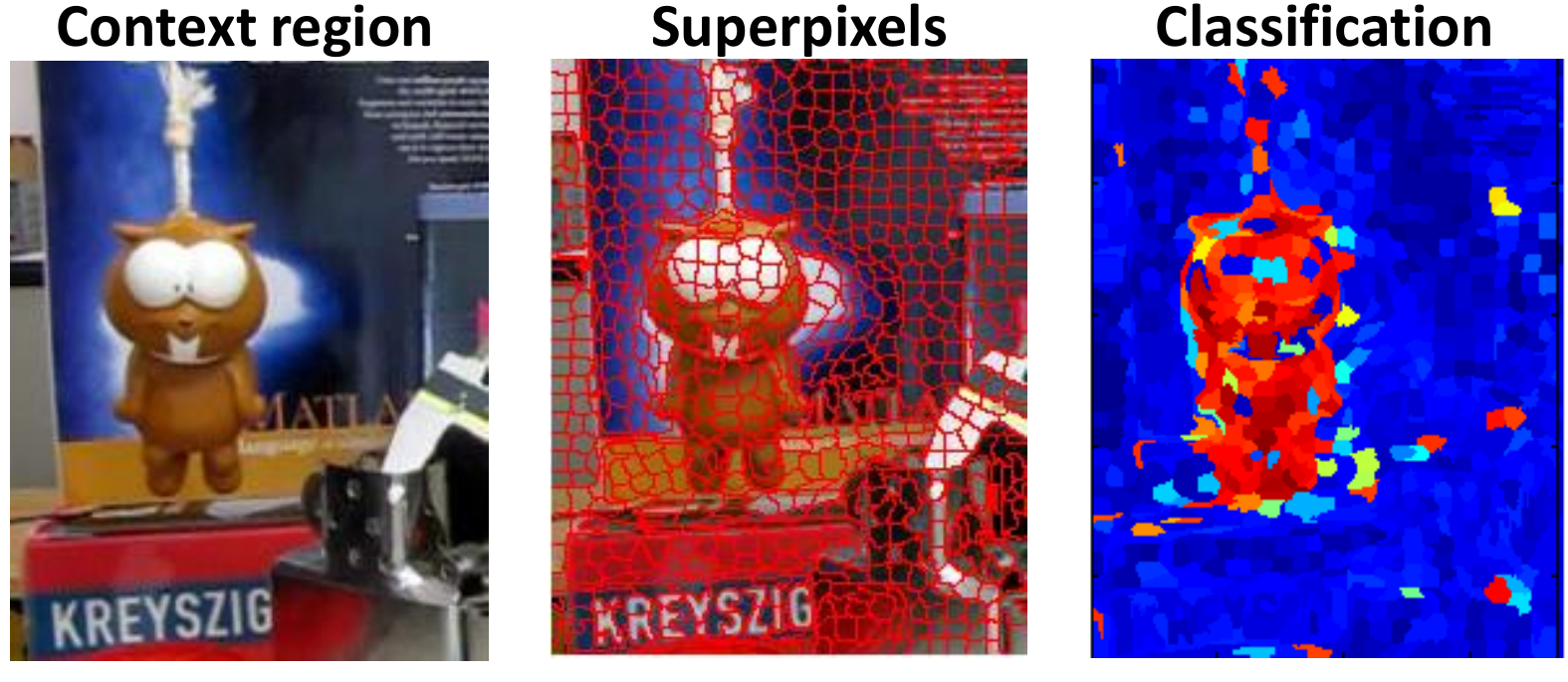}
\end{center}
\vspace{-0.33cm}
   \caption{Illustration of the local template-based visual representation using superpixels.}
    \label{fig:superpixel} \vspace{-0.33cm}
\end{figure*}

\item SIFT-based. Usually, a SIFT-based visual representation directly
makes use of the SIFT features inside an object region
to describe the structural information of object appearance.
Usually, there are two types of SIFT-based visual representations:
(i) individual SIFT point-based; and (ii) SIFT graph-based.
For (i), Zhou et al.~\citeyear{Zhou-Yuan-Shi-CVIU2009} set up a SIFT point-based
visual representation, and combine this visual representation with the mean shift
for object tracking. Specifically,
SIFT features are used to find the correspondences between the regions of interest
across frames. Meanwhile, the mean shift procedure is implemented to conduct a similarity
search via color histograms.  By using a mutual support mechanism between SIFT and the mean shift, the
tracking algorithm is able to achieve a consistent and stable tracking performance.  However, the
tracking algorithm may suffer from a background clutter which may lead to a one-to-many SIFT feature
matching. In this situation, the mean shift and SIFT feature matching may make mutually contradictory
decisions.
For (ii), the SIFT graph-based visual representations are based on
the underlying  geometric contextual relationship among SIFT feature points.
For example, Tang and Tao~\citeyear{Tang-Tao-TCSVT2008} construct a  relational graph  using SIFT-based attributes for object
representation.
The graph is based on the stable SIFT features which persistently appear in several consecutive
frames. However, such stable SIFT features are unlikely to exist in complex situations such as
shape deformation and illumination changes.

\item MSER-based. A MSER-based visual representation needs to
extract  the MSER (maximally stable extremal region) features for visual representation~\cite{Sivic-Schaffalitzky-Zisserman-ijcv2006}.
Subsequently, Tran and Davis~\citeyear{Tran-Davis-iccv2007}
construct a probabilistic pixel-wise occupancy map for each MSER feature,
and then perform the MSER feature matching for object tracking.
Similar to~\cite{Tran-Davis-iccv2007}, Donoser and Bischof~\citeyear{Donoser-Bischof-CVPR2006} also use
MSER features for visual representation. To improve the stability of MSER features, they
take temporal information across frames into consideration.

\item SURF-based. With the scale-invariant and rotation-invariant properties, the SURF (Speeded Up Robust Feature) is a variant of SIFT~\cite{Bay-Tuytelaars-Gool-ECCV2006}.
It  has similar
properties to those of SIFT in terms of repeatability, distinctiveness, and robustness, but its
computational speed is much faster.  Inspired by this fact, He et al.~\citeyear{He-Yamashita-Lu-Lao-ICCV2009} develop a
tracking algorithm using a SURF-based visual representation.  By
judging the compatibility of local SURF features with global object motion, the tracking algorithm
is robust to appearance changes and background clutters.

\item Corner feature-based. Typically, a corner feature-based visual representation makes use of corner features inside an object region
to describe the structural properties of object appearance,
and then
matches these corner features across frames for object localization.
For instance, Kim~\citeyear{Kim-CVPR2008} utilizes corner features for visual representation,
and then perform dynamic multi-level corner feature grouping to generate
a set of corner point trajectories. As a result, the spatio-temporal
characteristics of object appearance can be well captured.
Moreover, Grabner et al.~\citeyear{Grabner-Grabner-Bischof-cvpr2007}
explore the intrinsic differences between
the object and non-object corner features
by building a boosting discriminative model
for corner feature classification.

\item Local feature pool based. Recently, local feature pool based visual representations
have been widely used in ensemble learning based object tracking.
Usually, they need to set up a huge feature pool (i.e., a large number of various features) for
constructing a set of weak learners, which are used for discriminative feature selection.
Therefore, different kinds of visual features (e.g., color, local binary pattern~\cite{Collins-Liu-Leordeanu-PAMI2005},
histogram of oriented gradients~\cite{Collins-Liu-Leordeanu-PAMI2005,Liu-Yu-ICCV2007,Yu-Dinh-Medioni-ECCV2008},
Gabor features with Gabor wavelets~\cite{Nguyen-Smeulders-ECCV2004}, and Haar-like features with
Haar wavelets~\cite{Babenko-Yang-Belongie-CVPR2009}) can be used by FSSL in an independent or interleaving
manner. For example, Collins et al.~\citeyear{Collins-Liu-Leordeanu-PAMI2005}
set up a color feature pool whose elements are linear combinations of the following
RGB components:  $\{(\alpha_{1}, \beta_{1}, \gamma_{1})|\alpha_{1}, \beta_{1}, \gamma_{1} \in \{-2, -1, 0, 1, 2\}\}$.
As a result, an object is localized by selecting the discriminative color features from this pool.
Grabner and Bischof \cite{Grabner-Bischof-CVPR2006} construct an ensemble
classifier by learning several
weak classifiers trained from the Haar-like features~\cite{Viola-Jones-IJCV2002}, histograms of
oriented gradient (HOG)~\cite{Dalal-Triggs-CVPR2005}, and local binary patterns (LBP)
\cite{Ojala-Pietikainen-Maenpaa-PAMI2002}.
Babenko et al.~\citeyear{Babenko-Yang-Belongie-CVPR2009} utilize the Haar-like features
to construct a weak classifier, and then apply
an online multiple instance boosting
to learn a strong ensemble classifier for object tracking.

\item Saliency detection-based. In principle, saliency detection is
inspired by the focus-of-attention (FoA) theory~\cite{Palmer-MIT1999,Wolfe-1994} to
simulate the human perception mechanism for
capturing the salient information of an image. Such
salient information is helpful for visual object tracking due to
its distinctness and robustness.
Based on saliency detection, researchers
apply the biological vision theory to visual object tracking~\cite{Toyama-Hager-IJCV1996,Mahadevan-Vasconcelos-CVPR2009}.
More recently, Yang et al.~\citeyear{Yang-Yuan-Wu-cvpr2007,Fan-Wu-Dai-ECCV2010}
construct an attentional visual representation method based on the spatial selection.
This visual representation method takes a two-stage strategy for
spatial selective attention.  At the first stage, a pool of attentional regions (ARs) are
extracted as the salient image regions. At the second stage, discriminative learning is
performed to select several discriminative attentional regions for visual representation.
Finally, the task of object tracking is taken by matching the ARs between two consecutive
frames.
\end{itemize}

\subsubsection{Discussion}

    The aforementioned local feature-based representations use local templates, segmentation,
    SIFT, MSER, SURF, corner
    points, local feature pools, or saliency detection, respectively.  Due to the use of different features, these
    representations have different properties and characteristics. By representing an object region using a set of part templates,
    the local template-based visual representations are able to encode the
    local spatial layout information of object appearance, resulting in the robustness
    to partial occlusions. With the power of image segmentation,
    the segmentation-based visual representations are capable of well capturing the intrinsic structural
    information (e.g., object boundaries and superpixels) of object appearance, leading to
    reliable tracking results in challenging situations.
     Since the SIFT features are invariant to image scaling, partial
    occlusion, illumination change, and 3D camera viewpoint change,
    the SIFT-based representation is
    robust to appearance changes in illumination, shape deformation, and partial occlusion. However, it cannot
    encode precise information
    on the objects such as size, orientation, and pose.
    The MSER-based representation attempts to find several maximally stable
    extremal regions for feature matching across frames. Hence, it can tolerate pixel noise, but
    suffers from illumination changes.
    The SURF-based representation is on the basis of the ``Speeded Up Robust Features'', which has
    the properties of scale-invariance, rotation-invariance, and computationally efficiency.
     The corner-point representation aims to
    discover a set of corner features for feature matching. Therefore, it is suitable for tracking
    objects (e.g., cars or trucks) with plenty of corner points, and sensitive to the
    influence of non-rigid shape deformation and noise.  The feature pool-based representation is
    strongly correlated with feature selection-based ensemble learning that needs a number of local
    features (e.g., color, texture, and shape).  Due to the use of many features,  the process of feature extraction and feature selection is
    computationally slow.
The saliency detection-based representation aims to find a pool of
discriminative salient regions for a particular object.
By matching the salient regions across frames, object localization can be achieved.
However, its drawback is to rely heavily on
salient region detection, which is sensitive to noise or drastic illumination variation.

\subsection{Discussion on global and local visual representations}

    In general, the global visual representations are simple and
    computationally efficient for fast object tracking. Due to the imposed global geometric
    constraints, the global visual representations are susceptible to global appearance changes (e.g., caused by
    illumination variation or out-of-plane rotation). To deal with complicated appearance changes, a
    multi-cue strategy is taken by the global features to incorporate multiple types of visual
    information (e.g., position, shape, texture, and geometric structure) into the appearance
    models.

    In contrast, the local visual representations are able to capture
    the local structural object appearance. Consequently, the local visual representations are robust to
    global appearance changes caused by illumination variation, shape deformation, rotation, and
    partial occlusion. Since they require the keypoint detection, the interest point-based local visual representations often
    suffer from noise disturbance and background distraction.  Moreover, the local feature
    pool-based visual representations, which are typically required by discriminative feature selection, need a huge
    number of local features (e.g., color, texture, and shape), resulting in a very high
    computational cost.
    Inspired by the biological vision, the local visual representations using biological features
    attempt to capture the salient or intrinsic structural information inside the object regions.
    This salient information is relatively stable during the process of visual object tracking. However,
    salient region features rely heavily on salient region detection which may be susceptible to
    noise or drastic illumination variation, leading to potentially many feature mismatches across frames.

\section{Statistical modeling for tracking-by-detection}
\label{sec:statistical_modeling}

\begin{figure*}[t]
\begin{center}
\includegraphics[width=0.8\linewidth]{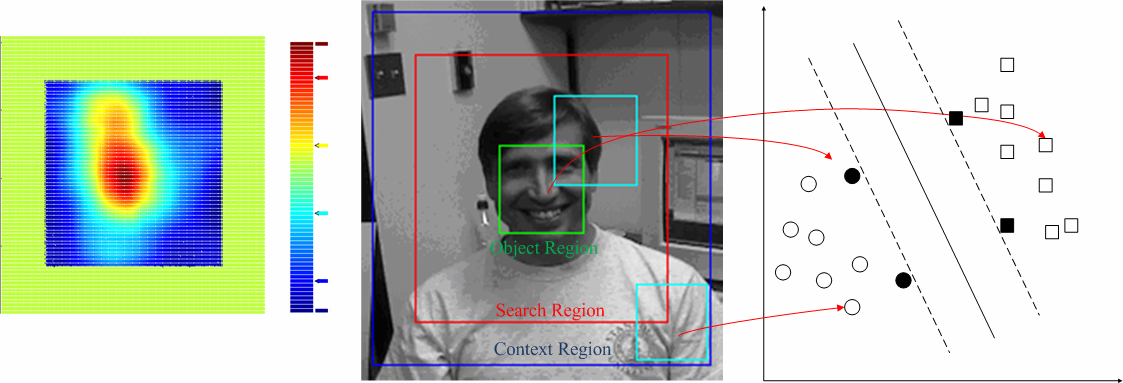}
\end{center}
\vspace{-0.33cm}
   \caption{Illustration of tracking-by-detection based on SVM classification (from [Tian et al. 2007], $\copyright$2007 Springer).
   The left subfigure shows the score map of face/non-face classification; the middle
   subfigure displays the search region for object localization and the context region
   for selecting face and non-face samples; the right subfigure plots the classification hyperplane
   that separates face and non-face classes.}
    \label{fig:svm_classification}
\end{figure*}

Recently, visual object tracking has been posed as a tracking-by-detection problem (shown in Fig.~\ref{fig:svm_classification}),
where statistical modeling is dynamically performed
to support object detection.
According to the model-construction mechanism,
statistical modeling is classified into
three categories, including generative, discriminative, and hybrid
generative-discriminative.

The generative appearance models mainly concentrate on how to accurately fit
the data from the object class.  However, it is very difficult to
verify the correctness of the specified model in practice.  Besides, the local optima are always
obtained during the course of parameter estimation (e.g., expectation maximization).  By
introducing online-update mechanisms, they incrementally learn visual representations for the
foreground object region information while ignoring the influence of the background. As a result,
they often suffer from distractions caused by the background regions with similar appearance to
the object class. Tab.~\ref{tab:generative_tracking} lists  representative
tracking-by-detection methods based on generative learning techniques.

In comparison, discriminative appearance models pose visual object tracking as a binary
classification issue.  They aim to maximize the separability between the object and non-object
regions discriminately.  Moreover, they focus on
discovering highly informative features for visual object tracking.  For the computational consideration,
online variants are proposed to incrementally learn discriminative classification
functions for the purpose of object or non-object predictions.  Thus, they can achieve effective
and efficient predictive performances.  Nevertheless, a major limitation of the discriminative
appearance models is to rely heavily on training sample selection (e.g., by self-learning or
co-learning).
Tab.~\ref{tab:discriminative_tracking} lists representative
tracking-by-detection methods based on discriminative learning techniques.

The generative and discriminative appearance models have their own advantages and disadvantages,
and are complementary to each other to a certain extent. Therefore,  researchers propose
hybrid generative-discriminative appearance models (HGDAMs) to fuse the useful information from
the generative and the discriminative models.  Due to taking a heuristic fusion strategy,
HGDAMs cannot guarantee that the performance of the hybrid models after information fusion is
better than those of the individual models. In addition, HGDAMs may add more constraints and
introduce more parameters, leading to more inflexibility in practice.
Tab.~\ref{tab:hybrid_tracking} lists representative
tracking-by-detection methods based on hybrid generative-discriminative learning techniques.

\begin{table*}[t]
\vspace{-0.45cm}
\caption{Summary of representative tracking-by-detection appearance models based on generative learning techniques.}
\label{tab:generative_tracking}
\hspace{-0.0cm}
\begin{center}
\scalebox{0.51}{
\begin{tabular}{ c|c||c|c|c||c }\hline
Item No. &
\begin{tabular}{c}
References\\
\end{tabular}
&
\makebox[0.8em]{
\begin{tabular}{c}
Mixture \\
models
\end{tabular}} &
\begin{tabular}{c}
Kernel density \\
estimation
\end{tabular}
&
\begin{tabular}{c}
Subspace \\
learning
\end{tabular}  &
\begin{tabular}{c}
Used
generative \\
learning techniques
\end{tabular}
 \\\hline\hline

 1&
\begin{tabular}{c}
\cite{McKenna-Raja-Gong-IVC1999}\\
\end{tabular}
&
\makebox[0.8em]{
\begin{tabular}{c}
color-based\\
GMM
\end{tabular}}
& --- & --- &
\begin{tabular}{c}
Gaussian mixture model (GMM)\\
in the hue-saturation
color space
\end{tabular}
\\ [0 ex] \hline

2&
\begin{tabular}{c}
\cite{Yu-Wu7}\\
\cite{Wang-Suter-Schindler-PAMI2007}\\
\end{tabular}
&
\begin{tabular}{c}
Spatio-color\\
GMM
\end{tabular}
 & --- & --- &
\begin{tabular}{c}
Spatial-color appearance model \\
using GMM\\
Spatial-color mixture of \\
Gaussians (SMOG)
\end{tabular}
\\ [0 ex] \hline

3&
\begin{tabular}{c}
\cite{Jepson-Fleet-Yacoob-PAMI2003,Zhou-Chellappa-Moghaddam6}\\
\end{tabular}
&
\begin{tabular}{c}
WSL
\end{tabular}
& --- & --- &
\begin{tabular}{c}
three-component mixture models:\\
W-component, S-component, L-component
\end{tabular}
\\ [0 ex] \hline

4&
\begin{tabular}{c}
\cite{Comaniciu-Ramesh-Meer-TPAMI}\\
\cite{Leichter-Lindenbaum-Rivlin-CVIU2010}\\
\end{tabular}
& --- &
\begin{tabular}{c}
Color-driven
\end{tabular}
& --- &
\begin{tabular}{c}
Mean shift using a spatially weighted \\
color histogram\\
Mean shift using multiple reference \\
color histograms
\end{tabular}
\\ [0 ex] \hline
5&
\begin{tabular}{c}
\cite{Leichter-Lindenbaum-Rivlin-PAMI2009}\\
\end{tabular}
& --- &
\begin{tabular}{c}
Shape-integration
\end{tabular}
& --- &
\begin{tabular}{c}
Affine kernel fitting\\
 using color and boundary cues
\end{tabular}
\\ [0 ex] \hline
6&
\begin{tabular}{c}
\cite{Collins-CVPR2003}\\
\cite{Yang-Duraiswami-Davis-CVPR2005}\\
\end{tabular}
& --- &
\begin{tabular}{c}
Scale-aware
\end{tabular}
& --- &
\begin{tabular}{c}
Mean shift considering scale changes\\
\end{tabular}
\\ [0 ex] \hline
7&
\begin{tabular}{c}
\cite{Nguyen-Robles-Kelly-Shen-cvpr2007}\\
\end{tabular}
& --- &
\begin{tabular}{c}
Scale-aware
\end{tabular}
& --- &
\begin{tabular}{c}
EM-based maximum likelihood \\
estimation
for kernel-based tracking
\end{tabular}
\\ [0 ex] \hline
8&
\begin{tabular}{c}
\cite{Alper-Yilmaz-cvpr2007}\\
\end{tabular}
& --- &
\begin{tabular}{c}
Non-symmetric\\
kernel
\end{tabular}
& --- &
\begin{tabular}{c}
Asymmetric kernel mean shift\\
\end{tabular}
\\ [0 ex] \hline
9&
\begin{tabular}{c}
\cite{Shen-Brooks-Hengel-TIP2007}\\
\end{tabular}
& --- &
\begin{tabular}{c}
Global \\
mode seeking\\
\end{tabular}
& --- &
\begin{tabular}{c}
Annealed mean shift\\
\end{tabular}
\\ [0 ex] \hline

10&
\begin{tabular}{c}
\cite{Han-Comaniciu-Zhu-Davis-PAMI2008}\\
\end{tabular}
& --- &
\begin{tabular}{c}
Sequential kernel\\
density estimation
\end{tabular}
& --- &
\begin{tabular}{c}
Sequential kernel-based tracking\\
\end{tabular}
\\ [0 ex] \hline

11&
\begin{tabular}{c}
\cite{Black-Jepson2,Ho-Lee-Yang}\\
\cite{Limy-Ross17,wen2012online}\\
\cite{wang2012object}
\end{tabular}
& --- & ---
&
\begin{tabular}{c}
Vector-based
linear \\
subspace learning
\end{tabular}
 &
\begin{tabular}{c}
Principal component analysis\\
Partial least square analysis\\
\end{tabular}
\\ [0 ex] \hline

12&
\begin{tabular}{c}
\cite{Wang-Gu-Shi-ICASSP2007,lixi-iccv2007}\\
\cite{Wen-Gao-Li-Tao-SMC2009,Hu-Li-IJCV2010}\\
\end{tabular}
& ---
& --- &
\begin{tabular}{c}
Tensor-based
linear \\
subspace learning
\end{tabular}&
\begin{tabular}{c}
2D principle component analysis\\
Tensor subspace analysis\\
\end{tabular}
\\ [0 ex] \hline

13&
\begin{tabular}{c}
\cite{Lim-Morariu-Camps-Sznaier15,Chin-Suter-TIP2007}\\
\end{tabular}
& ---
& --- &
\begin{tabular}{c}
Nonlinear \\
subspace learning
\end{tabular}&
\begin{tabular}{c}
Local linear embedding\\
Kernel principle component analysis\\
\end{tabular}
\\ [0 ex] \hline

14&
\begin{tabular}{c}
\cite{Meo-Ling-ICCV09,Li-Shen-Shi-cvpr2011}\\
\cite{zhang2012robust-multitask,jia2012visual}\\
\cite{bao2012real}\\
\end{tabular}
& ---
& --- &
\begin{tabular}{c}
Sparse representation
\end{tabular}&
\begin{tabular}{c}
$\ell_{1}$ sparse approximation
\end{tabular}
\\ [0 ex] \hline

15&
\begin{tabular}{c}
\cite{li2012non}\\
\end{tabular}
& ---
& --- &
\begin{tabular}{c}
Non-sparse representation
\end{tabular}&
\begin{tabular}{c}
Metric-weighted least-square regression
\end{tabular}
\\ [0 ex] \hline

16&
\begin{tabular}{c}
\cite{li2012incremental}\\
\end{tabular}
& ---
& --- &
\begin{tabular}{c}
3D-DCT representation
\end{tabular}&
\begin{tabular}{c}
Signal compression
\end{tabular}
\\ [0 ex] \hline

17&
\begin{tabular}{c}
\cite{Lee-Kriegman14,Fan-Yang-Wu-ICIP2008}\\
\cite{Kwon-Lee-CVPR2010}\\
\end{tabular}
& ---
& --- &
\begin{tabular}{c}
Multiple \\
subspaces
\end{tabular}&
\begin{tabular}{c}
bi-subspace or\\
multi-subspace learning
\end{tabular}
\\ [0 ex] \hline

18&
\begin{tabular}{c}
\cite{hou2001direct}\\
\cite{sclaroff2003active}\\
\cite{matthews2004active}\\
\end{tabular}
& ---
& --- &
\begin{tabular}{c}
Active appearance models
\end{tabular}&
\begin{tabular}{c}
Shape and appearance\\
3D mesh fitting
\end{tabular}
\\ [0 ex] \hline

\end{tabular}
}
\end{center}
\vspace{-0.2cm}
\end{table*}

\subsection{Mixture generative appearance models}

Typically, this type of generative appearance models
adaptively learns several components to capture the
spatio-temporal diversity of object appearance.
They can be classified into two categories:
WSL mixture models and Gaussian mixture models.

\begin{figure*}[t]
\begin{center}
\includegraphics[width=0.58\linewidth]{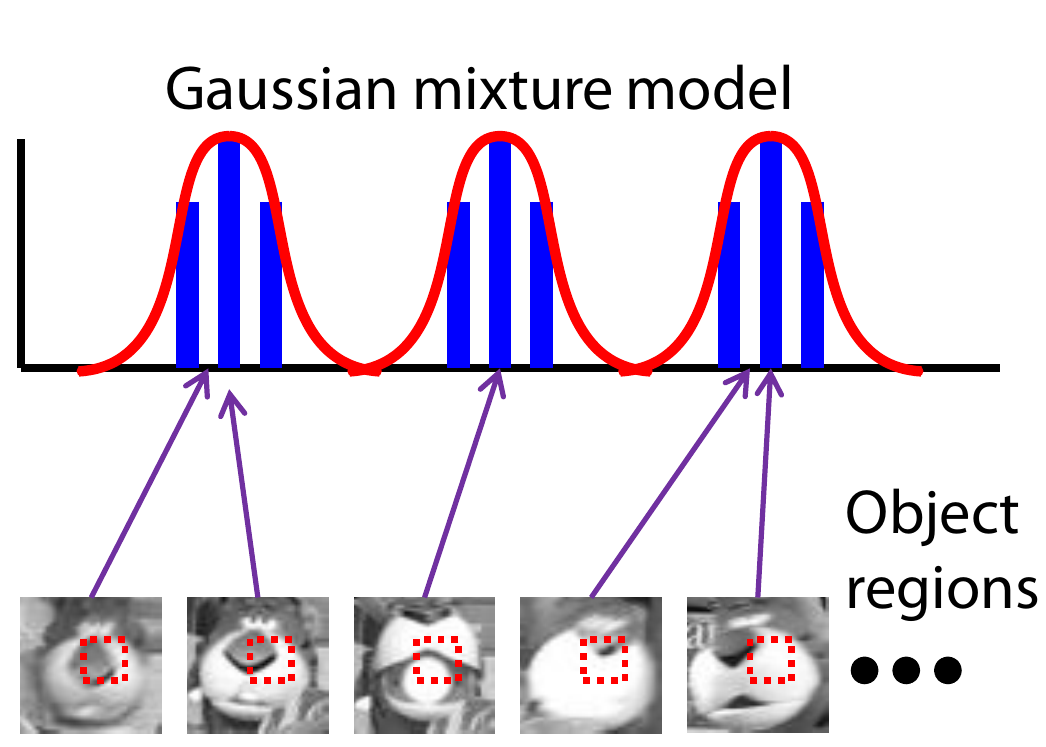}
\end{center}
\vspace{-0.33cm}
   \caption{Illustration of Gaussian mixture generative appearance models.}
    \label{fig:GMM} \vspace{-0.13cm}
\end{figure*}

\begin{itemize}
\item WSL mixture models. In principle, the WSL mixture model~\cite{Jepson-Fleet-Yacoob-PAMI2003}
contains the following three components: $W$-component, $S$-component, and $L$-component.
These three components characterize the inter-frame variations, the stable structure for all past
observations, and outliers such as occluded pixels, respectively.
As a variant of~\cite{Jepson-Fleet-Yacoob-PAMI2003}, another WSL mixture model~\cite{Zhou-Chellappa-Moghaddam6} is proposed
to directly
employ the pixel-wise intensities as visual features instead of using the
filter responses (e.g. in~\cite{Jepson-Fleet-Yacoob-PAMI2003}).
Moreover, the $L$-component is discarded in modeling the occlusion
using robust statistics, and an $F$-component is added as a fixed template that is observed most
often.

\item Gaussian mixture models. In essence, the Gaussian mixture models~\cite{McKenna-Raja-Gong-IVC1999,Stauffer-Grimson-PAMI2000,Han-Davis-ICCV2005,Yu-Wu7,Wang-Suter-Schindler-PAMI2007} utilize a set of
Gaussian distributions to approximate the underlying density function of object appearance, as shown in Fig.~\ref{fig:GMM}.
For instance, an
object appearance model~\cite{Han-Davis-ICCV2005} using a mixture of Gaussian density functions is proposed to
automatically determine the number of density functions and their
associated parameters including mean, covariance, and weight.
Rectangular features are introduced by averaging the corresponding intensities of
neighboring pixels (e.g., $3\times3$ or $5\times5$) in each color channel.
To capture a spatial-temporal description of the tracked objects,
Wang et al.~\citeyear{Wang-Suter-Schindler-PAMI2007} present a Spatial-color
Mixture of Gaussians (referred to as SMOG) appearance model,
which can simultaneously encode both spatial layout and color information.
To enhance its robustness and stability, Wang et al. further integrate multiple cues
into the SMOG appearance model, including three features of edge points:
their spatial distribution,  gradient intensity, and  size.
However, it is difficult for the Gaussian mixture models to select the correct number of components.
For example, adaptively determining the component number $k$ in a GMM is a difficult task in
practice.
As a result, the mixture models often use  ad-hoc or heuristic criteria for
selecting $k$, leading to the tracking inflexibility.
\end{itemize}

\subsection{Kernel-based generative appearance models (KGAMs)}

Kernel-based generative appearance models (KGAMs) utilize kernel density estimation
to construct kernel-based visual representations, and then carry out the mean shift for object localization, as shown in Fig.~\ref{fig:mean_shift}.
According to the mechanisms used for kernel construction or mode seeking, they may be split into
the following six branches: color-driven KGAMs, shape-integration KGAMs, scale-aware KGAMs, non-symmetric KG\-A\-Ms, KGAMs by global mode seeking,
and sequential-kernel-learning K\-G\-AMs.

\begin{itemize}
\item Color-driven KGAMs. Typically, a color-driven KGAM~\cite{Comaniciu-Ramesh-Meer-TPAMI} builds a color histogram-based
visual representation regularized by a spatially smooth isotropic kernel.
Using the Bhattacharyya coefficient as the similarity metric, a mean shift procedure is performed
for object localization by finding the basin of attraction of the local maxima.  However, the tracker~\cite{Comaniciu-Ramesh-Meer-TPAMI}
only considers color information and therefore ignores
other useful information such as edge and shape, resulting in the sensitivity to background clutters and
occlusions. Another color-driven KGAM~\cite{Leichter-Lindenbaum-Rivlin-CVIU2010}  is developed
to handle multi-view color variations by constructing
the convex hull of  multiple view-specific reference color histograms.

\item Shape-integration KGAMs. In general, the shape-integration KGAMs aim to
build a kernel density function in the joint color-shape space.
For example,
a shape-integration KGAM~\cite{Leichter-Lindenbaum-Rivlin-PAMI2009} is
proposed to capture the spatio-temporal properties of object appearance
using color and boundary cues.
It is based on two spatially normalized and rotationally symmetric kernels for
describing the information about the color and object
boundary.

\item Scale-aware KGAMs. In essence, the scale-aware KGAMs are to capture the spatio-temporal
distribution information on object appearance at multiple scales.
For instance, a scale-aware KGAM~\cite{Collins-CVPR2003} using the difference of Gaussian
based mean shift features is presented  to cope with the
problem of kernel scale selection by detecting
local maxima of the Difference-of-Gaussian (DOG) scale-space filters formulated as: 
\begin{equation}
    {\rm DOG}(x;\sigma)=\frac{1}{2\pi\sigma^{2}/1.6}\exp(-\frac{\|x\|^{2}}{2\sigma^{2}/1.6})-\frac{1}{2\pi\sigma^{2}(1.6)}\exp(-\frac{\|x\|^{2}}{2\sigma^{2}(1.6)})
\end{equation}
where $\sigma$ is a scaling factor.
Based on a
new probabilistic interpretation, another scale-aware KGAM~\cite{Nguyen-Robles-Kelly-Shen-cvpr2007} is proposed
to solve a maximum likelihood
problem, which treats the coordinates for the pixels as random variables.
As a result, the
problem of kernel scale selection is converted to that of
maximum likelihood optimization in the joint spatial-color space.

\item Non-symmetric KGAMs. The conventional KGAMs use a symmetric
kernel (e.g., a circle or an ellipse), leading to a large estimation bias
in the process of estimating the complicated underlying density function.
To address this issue, a non-symmetric KGAM~\cite{Alper-Yilmaz-cvpr2007} is developed
based on the asymmetric kernel mean shift with adaptively varying
the scale and orientation of the kernel.
In contrast to the symmetric mean shift (only requiring the image coordinate estimate),
the non-symmetric KGAM needs to simultaneously estimate
the image coordinates,
the scales, and the orientations in a few number of mean shift
iterations.
Introducing asymmetric kernels can generate a more accurate representation of the underlying
density so that the estimation bias is reduced.
Furthermore, the asymmetric kernel is just a generalization
of the previous radially symmetric
and anisotropic kernels.

\begin{figure*}[t]
\begin{center}
\includegraphics[width=0.7\linewidth]{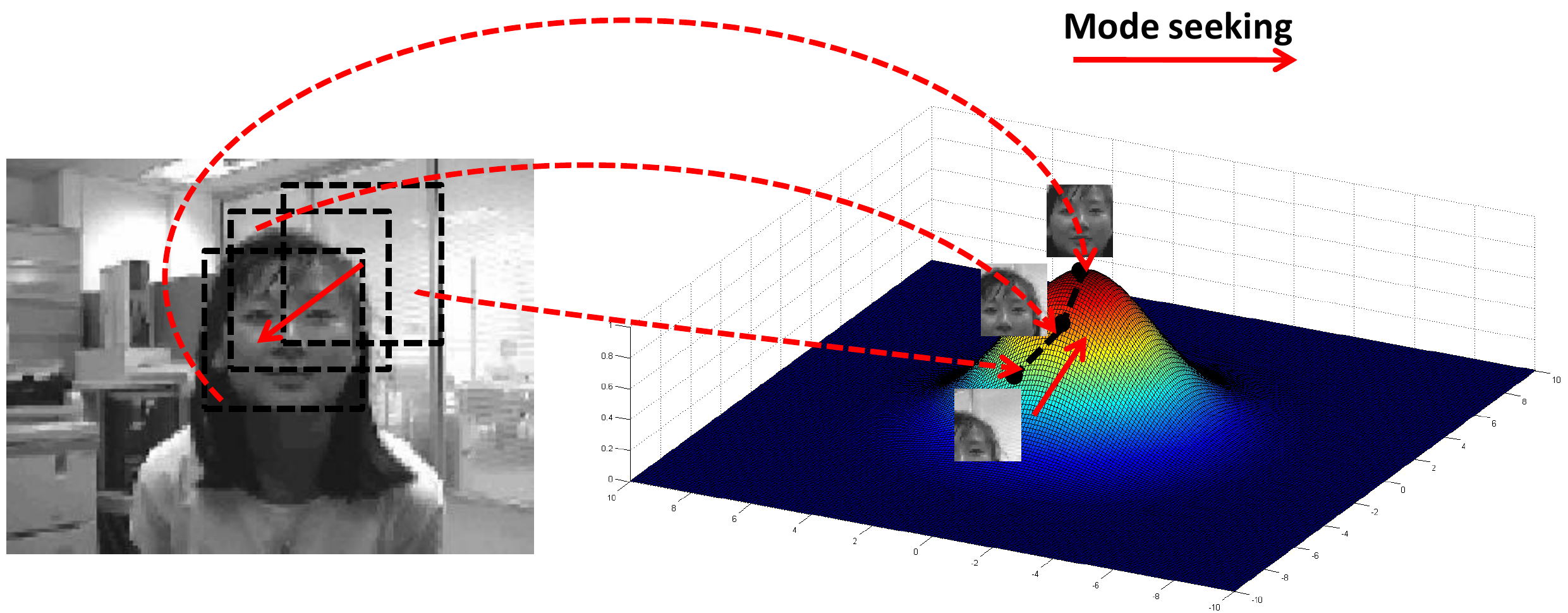}
\end{center}
   \caption{Illustration of the mode seeking process by mean shift.}
    \label{fig:mean_shift} 
\end{figure*}

\item KGAMs by global mode seeking. Due to the local optimization property of the mean shift, large inter-frame object translations
lead to tracking degradations or even failures.
In order to tackle this problem, Shen et al.~\citeyear{Shen-Brooks-Hengel-TIP2007}
propose an annealed mean shift algorithm motivated by the success of
the annealed importance sampling,
which is essentially a way of assigning the weights to the states obtained by multiple simulated
annealing runs~\cite{neal2001annealed}.
Here, the states correspond to the object positions while the simulated
annealing runs are associated with different bandwidths for the kernel density
estimation.
The proposed annealed mean shift algorithm
aims to make a progressive position evolution of the mean shift
as the bandwidths monotonically decrease (i.e., the convergence
position of mean shift with the last bandwidth works as the initial
position of the mean shift with the next bandwidth),
and finally seeks the global mode.

\item Sequential-kernel-learning KGAMs. Batch-mode kernel density estimation needs to store the nonparametric
representations of the kernel densities, leading to a high computational and memory complexity.
To address this issue,
Han et al.~\citeyear{Han-Comaniciu-Zhu-Davis-PAMI2008} develop a sequential kernel density approximation (SKDE)
algorithm for real-time visual object tracking.
The SKDE algorithm sequentially learns a
nonparametric representation of the kernel density and
propagates the density modes over time.

\item Discussion.
The color-driven kernel-based tracking algorithms mainly take the color information into consideration.
However, complicated factors may
give rise to drastic tracking degradations, including scale changes,
background clutters, occlusions, and rapid object movements.
To address this issue, various algorithmic extensions have been made.
The aim of scale-aware tracking algorithms is to capture the multi-scale
spatial layout information of object appearance. Thus, they are capable of
effectively completing the tracking task under the circumstance of
drastic scaling changes.
Moreover, the edge or shape information is very helpful for accurate object localization
or resisting background distraction.
Motivated by this consideration, shape-driven kernel-based tracking algorithms
have been developed to integrate the edge or shape information into the kernel
design process.
Normally, the kernel-based tracking algorithms utilize symmetric
kernels (e.g., a circle or an ellipse) for object tracking,
resulting in a large estimation bias
for complicated underlying density functions.
To tackle this problem, non-symmetric kernel-based tracking algorithms
are proposed to construct
a better representation of the underlying density.
Conventional
kernel-based tracking algorithms tend to pursue
the local model seeking, resulting in
tracking degradations or even failures due to their local optimization properties.
To address this issue, researchers borrow ideas from
both simulated annealing and
annealed importance sampling to obtain a feasible solution to global mode seeking.
In practice, the factors of computational complexity and memory consumption have
a great effect on real-time kernel-based tracking algorithms.
Thus, sequential techniques for kernel density estimation
have been developed for online kernel-based tracking.

\end{itemize}

\subsection{{Subspace learning-based generative appearance models (SLGAMs)}}

In visual object tracking, a target is usually associated with several underlying subspaces, each of which is spanned by a set of basis templates.
For convenience, let $\tau$ denote the target and $(\textbf{a}_{1} \hspace{0.1cm} \textbf{a}_{2} \ldots \textbf{a}_{N})$ denote
the basis templates of an underlying subspace.
Mathematically, the target $\tau$ can be linearly represented in the following form: 
\begin{equation}
\tau = c_{1}\textbf{a}_{1} + c_{2}\textbf{a}_{2} + \cdots + c_{N}\textbf{a}_{N} = (\textbf{a}_{1} \hspace{0.1cm} \textbf{a}_{2} \ldots \textbf{a}_{N})(c_{1} \hspace{0.1cm} c_{2} \ldots c_{N})^{T},
\end{equation}
where $(c_{1} \hspace{0.1cm} c_{2} \ldots c_{N})$ is the coefficient vector.
Therefore,  subspace learning-based generative appearance models (SLGAMs) focus on
how to effectively obtain these underlying subspaces and their associated
basis templates by using various techniques for subspace analysis.
For instance, some SLGAMs utilize eigenvalue decomposition or linear regression for subspace analysis, and
others construct multiple subspaces to model the distribution characteristics of
object appearance.
According to the used techniques for subspace analysis,
they can be categorized into two types: conventional and unconventional SLGAMs.

\begin{figure*}[t]
\begin{center}
\includegraphics[width=0.78\linewidth]{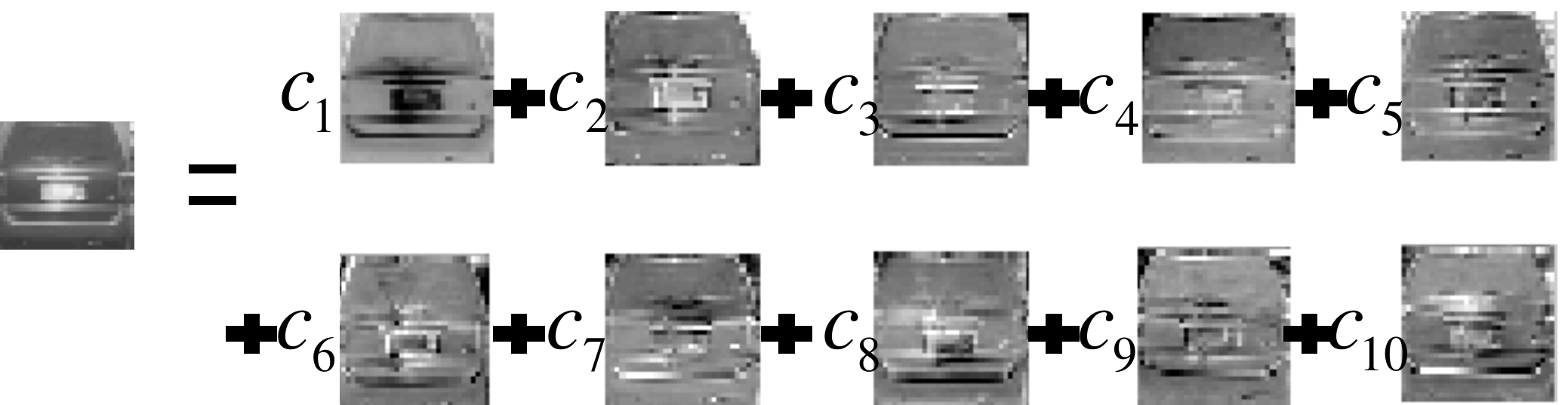}
\end{center}
\vspace{-0.33cm}
   \caption{Illustration of linear PCA subspace models. The left part shows
   a candidate sample, and the right part displays a linear combination
   of eigenbasis samples.}
    \label{fig:eigenbasis} \vspace{-0.13cm}
\end{figure*}

\subsubsection{{Conventional subspace models}}
In general, conventional subspace models can be split into the following two branches:
linear subspace models and non-linear subspace models.

\begin{itemize}
\item Linear subspace models. In recent years,
linear subspace models (LSMs) have been widely applied to visual object tracking.
According to the dimension of the used feature space, LSL can be divided into
(i) lower-order LSMs and (ii) higher-order LSMs.
The lower-order LSMs~\cite{Black-Jepson2,Ho-Lee-Yang,Li-Xu,Skocaj-Leonardis,wen2012online} needs to construct vector-based subspace models (e.g., eigenspace by principal component analysis shown in Fig.~\ref{fig:eigenbasis})
while the higher-order LSMs needs to build matrix-based or tensor-based subspace models
(e.g., 2D eigenspace by 2D principal component analysis and tensor eigenspace by tensor analysis).

\hspace{0.18cm} For (i), several incremental principal component analysis (PCA)
algorithms are proposed to make linear subspace models more efficient.
For instance, an incremental robust PCA algorithm~\cite{Li-Xu} is developed to
incorporate robust analysis into the process of subspace learning.
Similar to~\cite{Li-Xu}, Skocaj and Leonardis~\citeyear{Skocaj-Leonardis}
embed the robust analysis technique into the incremental
subspace learning framework,
which makes a sequential update of
the principal subspace.
The learning framework considers the weighted influence of both
individual images and individual pixels within an image.
Unlike the aforementioned robust PCA algorithm based on weighted residual errors,
the incremental subspace learning algorithms in \cite{Levy-Lindenbaum-TIP2000,Brand-ECCV2002}
utilize incremental singular value decomposition (SVD) to obtain a closed-form solution to
subspace learning. However, these incremental PCA algorithms cannot update the sample mean
during subspace learning. To address this issue,
a subspace model based on R-SVD (i.e., rank-R singular value decomposition) is built
with a sample mean update~\cite{Limy-Ross17}.
Moreover, Wang et al.~\citeyear{wang2012object} apply partial least square analysis to learn a low-dimensional
feature subspace for object tracking.
In theory, the partial least square analysis
is capable of modeling relations between sets of
variables driven by a small number of latent factors,
leading to robust object tracking results.

\hspace{0.18cm} For (ii), a set of higher-order LSMs are proposed to
address the small-sample-size problem, where the number of samples is far smaller
than the dimension of samples. Therefore, many researchers begin to build
matrix-based or tensor-based subspace models. For instance,
Wang et al.~\citeyear{Wang-Gu-Shi-ICASSP2007}
directly analyze the 2D image matrices, and
construct a 2DPCA-based appearance model for object tracking.
In addition to the foreground information, they also consider background information to avoid the
distractions from the background clutters.
Moreover, Li et al.~\citeyear{lixi-iccv2007,Hu-Li-IJCV2010} and Wen et al.~\citeyear{Wen-Gao-Li-Tao-SMC2009}
take advantage of
online tensor decomposition to construct a tensor-based appearance model for
robust visual object tracking.

\item Nonlinear subspace models.
If the training data lie on an underlying nonlinear manifold,
the LSM-based tracking algorithms may fail.
Therefore, researchers attempt to employ nonlinear subspace learning
to capture the underlying geometric information from target samples.
For the robust human tracking, a nonlinear subspace model~\cite{Lim-Morariu-Camps-Sznaier15} is
built using nonlinear
dimension reduction techniques (i.e., Local Linear Embedding).
As a nonlinear generalization of PCA, a nonlinear subspace model~\cite{Chin-Suter-TIP2007}
based on kernel principal component analysis (KPCA) is constructed
to capture the kernelized eigenspace information from target samples.
\end{itemize}

\subsubsection{{Unconventional subspace models}}
In general, unconventional subspace models
can also be used for visual object tracking. Roughly, they can be divided into
three categories: sparse/non-sparse representation, autoregressive modeling,
and multi-subspace learning.

\begin{itemize}
\item Sparse/non-sparse representation. Typically, a set of target samples is associated
with  an underlying subspace spanned by several templates.
The likelihood of a candidate sample belonging to the object class is often
determined by the residual between the candidate samples and the
reconstructed samples derived from a linear representation.
To ensure a sparse linear representation,  an
$\ell_{1}$-regularized optimization procedure
is adopted to obtain a sparse linear representation solution~\cite{Meo-Ling-ICCV09}.
Based on the sparse representation technique in~\cite{Meo-Ling-ICCV09},
Jia et al.~\citeyear{jia2012visual} propose a tracking method that further improves
the tracking accuracy by using the block-division spatial pooling schemes (e.g.,
average pooling, max pooling, and alignment pooling).
Moreover, Zhang et al.~\citeyear{zhang2012robust-multitask} present a multi-task
sparse optimization framework based on a $\ell_{p, q}$-regularized least-square
minimization cost function. Instead of treating test samples
independently, the framework explores the interdependencies between test samples
by solving a $\ell_{p, q}$-regularized group sparsity problem.
When $p=q=1$, the framework degenerates to the popular $\ell_{1}$ tracker~\cite{Meo-Ling-ICCV09}.

\hspace{0.18cm}To achieve a real-time performance of the $\ell_{1}$ tracker~\cite{Meo-Ling-ICCV09}, a subspace model~\cite{Li-Shen-Shi-cvpr2011} based on compressive sensing
is built by solving an orthogonal matching pursuit (OMP) optimization problem (i.e., random projections),
which is about 6000 times faster than~\cite{Meo-Ling-ICCV09}.
Similar to~\cite{Li-Shen-Shi-cvpr2011}, Zhang et al.~\citeyear{zhangCompressiveTracking2012} make use
of compressive sensing (random projections) to generate a low-dimensional compressive feature descriptor,
leading to a real-time tracking performance.
Alternatively, Bao et al.~\citeyear{bao2012real} take advantage of the popular accelerated proximal gradient (APG) approach
to optimize the $\ell_{1}$-regularized least square minimization problem, which has a
quadratic convergence property to ensure the real-time tracking performance.
Another way of improving the efficiency of the $\ell_{1}$ tracker~\cite{Meo-Ling-ICCV09}
is to reduce the number of $\ell_{1}$ minimizations in the process of evaluating test samples~\cite{mei2011minimum}.
This task is accomplished by estimating the minimal error bound of the likelihood function in particle filtering,
resulting in a moderate improvement in tracking efficiency.
From an viewpoint of signal compression, Li et al.~\cite{li2012incremental} construct
a compact 3D-DCT object representation based on a DCT subspace
spanned by cosine basis functions. With the power of fast Fourier Transform (FFT),
the proposed 3D-DCT object representation is capable of efficiently adapting
to spatio-temporal appearance variations during tracking, leading to
robust tracking results in complicated situations.

\hspace{0.18cm} On the other hand, the sparsity of the linear representation is
unnecessary for robust object tracking as long as an adequate number of
template samples are provided, as pointed out in~\cite{li2012non}. Therefore, a non-sparse metric weighted
linear representation (with a closed-form solution) is proposed
to effectively and efficiently model the intrinsic appearance
properties of the tracked object~\cite{li2012non}.

\item Autoregressive modeling. Since tracking is a time-dependent process, the
target samples from adjacent frames are mutually correlated. To characterize
the time dependency across frames, a variety of appearance models are
proposed in recent years. For instance, a dynamical statistical
shape representation is proposed to capture the
temporal correlation information on human silhouettes
from consecutive frames~\cite{Cremers-PAMI2006}. The proposed representation
learns a linear
autoregressive shape model, where the current silhouette
is linearly constrained by the previous silhouettes.
The learned shape model is then integrated into
the level-set evolution process, resulting in
robust segmentation results.

\begin{figure*}[t]
\vspace{-0.15cm}
\begin{center}
\includegraphics[width=0.68\linewidth]{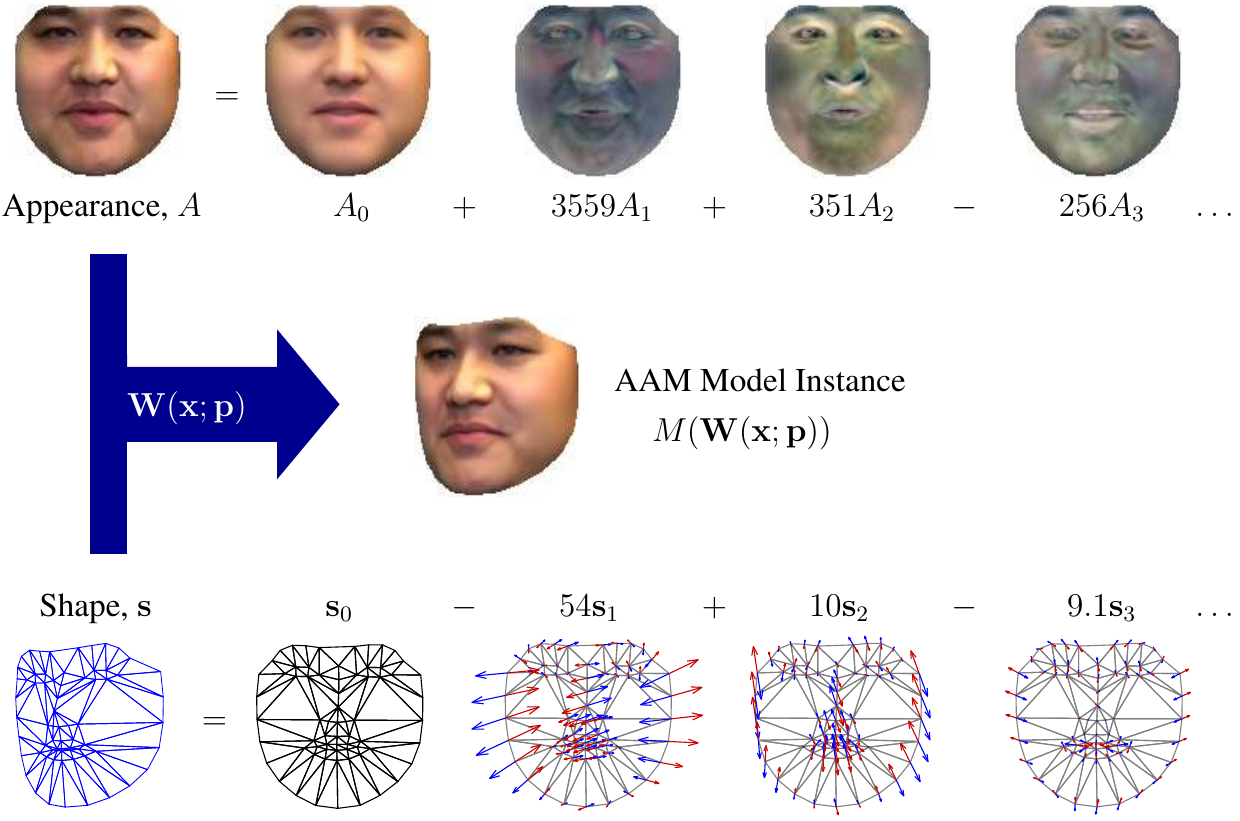}
\end{center}
\vspace{-0.33cm}
   \caption{Illustration of active appearance models (from [Matthews and Baker 2004], $\copyright$2004 Springer). The upper part shows that an appearance $A$
   is linearly represented by a base appearance $A_{0}$ and several appearance images;
   the middle part displays the piecewise affine warp $\textbf{W}(x; \textbf{p})$ that transforms a pixel from a base shape
   into the active appearance model; and the lower part exhibits that a shape s is linearly
   represented by a base shape $\mbox{s}_{0}$ and several shapes $(\mbox{s}_{i})_{i=1}^{n}$.}
    \label{fig:aam} \vspace{-0.1cm}
\end{figure*}

\item Multi-subspace learning.  In order to capture the distribution diversity of
target samples,
several efforts establish the double or multiple subspaces for
visual representation. For example,
Fan et al.~\citeyear{Fan-Yang-Wu-ICIP2008} present a bi-subspace model for visual
tracking. The model simultaneously considers two visual cues:
color appearance and texture appearance. Subsequently, the model uses a co-training strategy to
exchange information between two visual cues.
For video-based recognition and tracking, Lee and Kriegman~\citeyear{Lee-Kriegman14}
present a generic appearance model that seeks to
set up a face appearance manifold consisting of several sub-manifolds.
Each sub-manifold corresponds to a face pose subspace.
Furthermore, Kwon and Lee~\citeyear{Kwon-Lee-CVPR2010} construct a set of basic observation
models, each of which is associated with a specific appearance manifold of a tracked object. By
combining these basic observation models, a compound observation model is obtained, resulting in a
robustness to combinatorial appearance changes.

\item Active appearance models (AAMs). Usually, AAMs~\cite{hou2001direct,sclaroff2003active,matthews2004active} need to incorporate two
components: a) shape and b) appearance, as shown in Fig.~\ref{fig:aam}. For a),
the shape s of an AAM can be expressed as a linear combination of a base shape $\mbox{s}_{0}$ and several shape vectors $(\mbox{s}_{i})_{i=1}^{n}$ such that
$\mbox{s} = \mbox{s}_{0} + \sum_{i=1}^{n}p_{i}\mbox{s}_{i}$ where the shape s
denotes $(x_{1}, y_{1}, x_{2}, y_{2}, \ldots, x_{v}, y_{v})$ that
are the coordinates of the $v$ vertices making up the mesh.
For b), the appearance of the AAM can be represented as a linear combination of
a base appearance $A_{0}(x)$ and several appearance images $(A_{i}(x))_{i=1}^{m}$ such that
$A(x) = A_{0}(x) + \sum_{i=1}^{m}\lambda_{i}A_{i}(x)$ where
$x \in \mbox{s}_{0}$ is a pixel lying inside the base mesh $\mbox{s}_{0}$.
Therefore, given a test image, the AAM needs to minimize the following cost function
for the model fitting: \vspace{-0.19cm}
\begin{equation}
\sum_{x\in \mbox{s}_{0}}\left[A_{0}(x) + \sum_{i=1}^{m}\lambda_{i}A_{i}(x) - I(\textbf{W}(x; \textbf{p}))\right],
\vspace{-0.1cm}
\end{equation}
where $\textbf{W}(x; \textbf{p})$ denotes a piecewise affine warp that transforms a pixel $x \in \mbox{s}_{0}$
into AAM.
\end{itemize}

\vspace{-0.15cm}
\subsubsection{{Discussion}}
\vspace{-0.15cm}

The lower-order linear subspace models (LSMs) usually learn vector-based
visual representations for visual object tracking. For the tracking efficiency, several
incremental LSMs (e.g., incremental PCA) are developed for online visual object tracking.
Since the vector-based visual representations suffer from the small-sample-size problem,
researchers construct higher-order matrix-based or tensor-based visual representations.
However, the above LSMs potentially assume that object appearance samples
lie on an underlying linear manifold. In practice, this assumption is
often violated because of complex extrinsic/intrinsic appearance changes.
Motivated by this consideration, non-linear subspace models
are developed for visual representation.
However, the problem with these non-linear subspace models is that they are
computationally expensive due to the non-linear subspace learning (e.g., nonlinear dimension reduction).

In recent years,  unconventional subspace models
have been proposed for visual object tracking. These models
either enforce the sparsity constraints on the linear representation solution
or have different assumptions of subspace properties.
However, the sparsity-constrained linear representation typically
induces a high optimization complexity, which motivates researchers
to develop an efficient optimization method (e.g., APG and OMP) for a real-time tracking performance.
Without the conventional single-subspace assumption,
bi-subspace or multi-subspace algorithms are proposed
to more precisely model the distribution diversity of the target samples,
but at the cost of an additional computation.

\begin{table*}[t]
\vspace{-0.3cm}
\caption{{Summary of representative tracking-by-detection methods based on discriminative learning
techniques \vspace{-0.0cm}}}
\label{tab:discriminative_tracking}
\hspace{-0.0cm}
\scalebox{0.49}{
\begin{tabular}{c|c||c|c|c|c||c|c}\hline
Item No. &
\makebox[7em]{
\begin{tabular}{c}
References\\
\end{tabular}}
&
\makebox[5.2em]{
\begin{tabular}{c}
Boosting
\end{tabular}} &
\begin{tabular}{c}
SVM
\end{tabular}
&
\begin{tabular}{c}
Randomized \\
learning
\end{tabular}  &
\begin{tabular}{c}
Discriminant \\
analysis
\end{tabular}
&
\makebox[3.8em]{
\begin{tabular}{c}
Codebook \\
learning
\end{tabular}}
&
\begin{tabular}{c}
Used Discriminative\\
learning techniques
\end{tabular}
 \\\hline\hline

1&
\begin{tabular}{c}
\cite{Grabner-Grabner-Bischof-BMVC2006}\\
\cite{Grabner-Bischof-CVPR2006}\\
\cite{Liu-Yu-ICCV2007}\\
\end{tabular}
&
\makebox[3.8em]{
\begin{tabular}{c}
Self-learning\\
single-instance
\end{tabular}}
& --- & --- & --- & ---&
\begin{tabular}{c}
Boosting with
feature\\
 ranking-based
feature selection
\end{tabular}
\\ [0 ex] \hline

2&
\begin{tabular}{c}
\cite{Avidan-2007}\\
\end{tabular}
&
\makebox[3.8em]{
\begin{tabular}{c}
Self-learning\\
single-instance
\end{tabular}}
& --- & --- & --- & ---&
\begin{tabular}{c}
Boosting with
feature \\
weighting-based
feature selection
\end{tabular}
\\ [0 ex] \hline

3&
\begin{tabular}{c}
\cite{Visentini-Snidaro-Foresti-VS2008}\\
\end{tabular}
&
\makebox[3.8em]{
\begin{tabular}{c}
Self-learning\\
single-instance
\end{tabular}}
& --- & --- & --- & ---&
\begin{tabular}{c}
Dynamic ensemble \\
based
boosting\\
\end{tabular}
\\ [0 ex] \hline

4&
\begin{tabular}{c}
\cite{Leistner-Saffari-Roth-Bischof-OLCV2009}\\
\end{tabular}
&
\makebox[3.8em]{
\begin{tabular}{c}
Self-learning\\
single-instance
\end{tabular}}
& --- & --- & --- & ---&
\begin{tabular}{c}
Noise-insensitive \\
boosting\\
\end{tabular}
\\ [0 ex] \hline

5&
\begin{tabular}{c}
\cite{Okuma-Taleghani-Freitas-Little-Lowe-ECCV2004,Wang-Chen-Gao2005}\\
\end{tabular}
&
\makebox[3.8em]{
\begin{tabular}{c}
Self-learning\\
single-instance
\end{tabular}}
& --- & --- & --- & ---&
\begin{tabular}{c}
Particle filtering integration \\
based boosting\\
\end{tabular}
\\ [0 ex] \hline

6&
\begin{tabular}{c}
\cite{wu2012new,luo2011robust}\\
\end{tabular}
&
\makebox[3.8em]{
\begin{tabular}{c}
Self-learning\\
single-instance
\end{tabular}}
& --- & --- & --- & ---&
\begin{tabular}{c}
Transfer learning \\
based boosting\\
\end{tabular}
\\ [0 ex] \hline

7&
\begin{tabular}{c}
\cite{Levin-Viola-Freund-ICCV2007,Grabner-Grabner-Bischof-ECCV2008}\\
\cite{Liu-Cheng-Lu-ICCV2009}\\
\end{tabular}
&
\makebox[3.8em]{
\begin{tabular}{c}
Co-learning\\
single-instance
\end{tabular}}
& --- & --- & --- & ---&
\begin{tabular}{c}
Semi-supervised \\
co-learning boosting\\
\end{tabular}
\\ [0 ex] \hline

8&
\begin{tabular}{c}
\cite{Babenko-Yang-Belongie-CVPR2009,Li-Kwok-Lu-cvpr2010}\\
\end{tabular}
&
\makebox[3.8em]{
\begin{tabular}{c}
Self-learning\\
Multi-instance
\end{tabular}}
& --- & --- & --- & ---&
\begin{tabular}{c}
Multiple instance \\
boosting\\
\end{tabular}
\\ [0 ex] \hline

9&
\begin{tabular}{c}
\cite{Zeisl-Leistner-Saffari-Bischof-CVPR2010}\\
\end{tabular}
&
\makebox[3.8em]{
\begin{tabular}{c}
Co-learning\\
Multi-instance
\end{tabular}}
& --- & --- & --- & ---&
\begin{tabular}{c}
Semi-supervised\\
Multiple instance \\
boosting\\
\end{tabular}
\\ [0 ex] \hline

10&
\begin{tabular}{c}
\cite{Avidan-PAMI2004,Williams-Blake-Cipolla-PAMI2005}\\
\cite{Tian-Zhang-Liu-ACCV2007}\\
\end{tabular}
& ---
&
\makebox[4.9em]{
\begin{tabular}{c}
Self-learning\\
single-instance
\end{tabular}}
 & --- & --- & ---&
\begin{tabular}{c}
Single
SVM classifier\\
or SVM ensemble classifiers
\end{tabular}
\\ [0 ex] \hline

11&
\begin{tabular}{c}
\cite{bai2012robust}\\
\end{tabular}
& ---
&
\makebox[4.9em]{
\begin{tabular}{c}
Self-learning\\
single-instance
\end{tabular}}
 & --- & --- & ---&
\begin{tabular}{c}
Ranking SVM learning
\end{tabular}
\\ [0 ex] \hline

12&
\begin{tabular}{c}
\cite{hare2011struck_tracking,yao2012robust}\\
\end{tabular}
& ---
&
\makebox[4.9em]{
\begin{tabular}{c}
Self-learning\\
single-instance
\end{tabular}}
 & --- & --- & ---&
\begin{tabular}{c}
Structured SVM learning
\end{tabular}
\\ [0 ex] \hline

13&
\begin{tabular}{c}
\cite{Tang-Brennan-Zhao-ICCV2007}\\
\end{tabular}
& ---
&
\makebox[4.9em]{
\begin{tabular}{c}
Co-learning\\
single-instance
\end{tabular}}
 & --- & --- & ---&
\begin{tabular}{c}
Semi-supervised\\
SVM classifiers\\
\end{tabular}
\\ [0 ex] \hline

14&
\begin{tabular}{c}
\cite{Saffari-Leistner-Santner-Godec-Bischof-OLCV2009,Godec-Leistner-Saffari-Bischof-ICPR2010}\\
\cite{Leistner-Saffari-Bischof-ECCV2010}\\
\end{tabular}
 & ---
& ---
&
\makebox[4.9em]{
\begin{tabular}{c}
Self-learning\\
single-instance
\end{tabular}}
 & --- & ---&
\begin{tabular}{c}
Random forests or
Random \\
Naive Bayes classifiers\\
\end{tabular}
\\ [0 ex] \hline

15&
\begin{tabular}{c}
\cite{Lin-Yang-Levinson-ICPR2004,Nguyen-Smeulders-IJCV2006}\\
\cite{Li-Liang-Huang-Jiang-Gao-ICIP2008}\\
\end{tabular}
 & ---
& ---
&
\makebox[4.9em]{
\begin{tabular}{c}
Single-modal\\
self-learning\\
single-instance
\end{tabular}}
 & --- & ---&
\begin{tabular}{c}
Fisher Linear \\
Discriminant Analysis
\end{tabular}
\\ [0 ex] \hline

16&
\begin{tabular}{c}
\cite{wang2010discriminative,jiang2011tracking}\\
\cite{jiang2012order}\\
\end{tabular}
 & ---
& ---
&
\makebox[4.9em]{
\begin{tabular}{c}
Single-modal\\
self-learning\\
single-instance
\end{tabular}}
 & --- & ---&
\begin{tabular}{c}
Discriminant \\
metric learning
\end{tabular}
\\ [0 ex] \hline

17&
\begin{tabular}{c}
\cite{Zhu-Martinez-PAMI2006,Xu-Shi-Xu-2008}\\
\end{tabular}
 & ---
& ---
 & ---
&
\makebox[4.9em]{
\begin{tabular}{c}
Multi-modal\\
self-learning\\
single-instance
\end{tabular}}
& ---&
\begin{tabular}{c}
Subclass\\
Discriminant Analysis
\end{tabular}
\\ [0 ex] \hline

18&
\begin{tabular}{c}
\cite{Zhang-Hu-Maybank-Li-ICCV2007,Zha-Yang-Bi-PR2010}\\
\end{tabular}
 & ---
& ---
 & ---
&
\makebox[4.9em]{
\begin{tabular}{c}
Graph-driven\\
self-learning\\
single-instance
\end{tabular}}
& ---&
\begin{tabular}{c}
Graph embedding\\
Graph transductive learning
\end{tabular}
\\ [0 ex] \hline

19&
\begin{tabular}{c}
\cite{Collins-Liu-Leordeanu-PAMI2005}\\
\end{tabular}
 & ---
& ---
 & ---
 & ---
&
\makebox[4.9em]{
\begin{tabular}{c}
Self-learning\\
single-instance
\end{tabular}}
&
\begin{tabular}{c}
Feature ranking\\
based feature selection
\end{tabular}
\\ [0 ex] \hline

20&
\begin{tabular}{c}
\cite{Gall-Razavi-Gool-BMVC2010}\\
\end{tabular}
 & ---
& ---
 & ---
 & ---
&
\makebox[5.5em]{
\begin{tabular}{c}
Instance-specific\\
codebook
\end{tabular}}
&
\begin{tabular}{c}
Discriminative\\
codebook learning
\end{tabular}
\\ [0 ex] \hline

\end{tabular}
}
\end{table*}

\vspace{-0.1cm}
\subsection{Boosting-based discriminative appearance models}
\vspace{-0.1cm}

In the last decade,  boosting-based discriminative appearance models (BDAMs) have been widely used in
visual object tracking because of their powerful discriminative learning capabilities.  According to the
learning strategies employed, they can be categorized into self-learning and co-learning BDAMs.
Typically, the self-learning BDAMs utilize the discriminative information from single source to
guide the task of object/non-object classification, while the co-learning
BDAMs exploit the multi-source discriminative information
for object detection. More specifically,
the self-learning BDAMs first train a classifier over the data from the previous frames,
and subsequently use the trained classifier to evaluate possible object regions at the current frame.
After object localization, a set of so-called ``positive''
and ``negative'' samples are selected to update the classifier.
These ``positive''
and ``negative'' samples are
labeled by the previously trained classifier.
Due to tracking errors,
the training samples obtained in the tracking process may be polluted by noise.
Therefore, the labels for the training samples are unreliable.
As the tracking process proceeds, the tracking error may be accumulated,
possibly resulting in the ``drift'' problem.
In contrast, the co-learning BDAMs often takes a semi-supervised strategy for
object/non-object classification (e.g., co-training by building multiple classifiers).

On the
other hand, BDAMs also take different strategies for visual representation, i.e., single-instance
and multi-instance ones.
The single-instance BDAMs
require precise object localization.
If a precise object localization is not available, these tracking algorithms may use sub-optimal positive
samples to update their corresponding object or non-object discriminative
classifiers, which may lead to a model drift problem.
Moreover, object detection or tracking has its own inherent ambiguity,
that is, precise object locations may be unknown even for human labelers.
To deal with this ambiguity, the multi-instance BDAMs are proposed to represent
an object by a set of image patches around the tracker location.
Thus, they can be further classified into single-instance or multi-instance
BDAMs.

\subsubsection{{Self-learning single-instance BDAMs}}

Based on online boosting~\cite{Oza-Russell-AIS2001},
researchers have developed
a variety of computer vision applications such as
object detection~\cite{Viola-Jones-IJCV2002} and
visual object tracking~\cite{Grabner-Grabner-Bischof-BMVC2006,Grabner-Bischof-CVPR2006}.
In these applications, the variants of boosting are invented
to satisfy different demands.

\begin{figure*}[t]
\begin{center}
\includegraphics[width=0.73\linewidth]{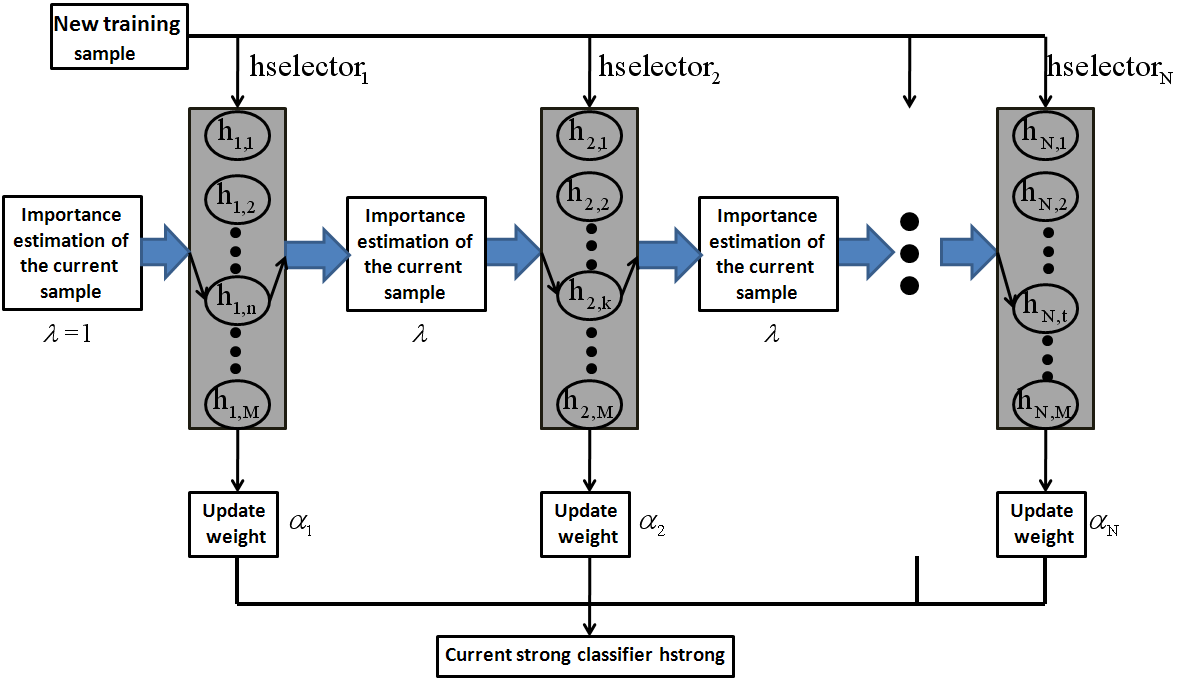}
\end{center}
\vspace{-0.3cm}
\caption{Illustration of online boosting for feature selection.}
\label{fig:online_boosting}
\end{figure*}

\begin{itemize}

\item Conventional BDAMs. As shown in Fig.~\ref{fig:online_boosting}, the conventional BDAMs first make a discriminative evaluation of each feature from a candidate
feature pool, and then select the top-ranked features to conduct the tracking process~\cite{Grabner-Grabner-Bischof-BMVC2006,Grabner-Bischof-CVPR2006}.
To accelerate the feature selection process,
Liu and Yu~\citeyear{Liu-Yu-ICCV2007} utilize
gradient-based feature selection to construct a BDAM. But this BDAM
requires an initial set of weak classifiers
to be given in advance,
leading to difficulty in general object tracking.
The above-mentioned BDAMs often perform poorly in
capturing the correlation information between features, leading to the redundancy
of selected features and the failure to compensate for the tracking error caused by other features.

To address this issue, a feature weighting strategy is adopted to attach
all the features from the feature pool with different weights,
and then performs weighted fusion for object tracking.
For instance,
Avidan~\citeyear{Avidan-2007}
constructs a confidence map by pixel classification using an
ensemble of online learned weak classifiers, which are trained by a
feature weighting-based boosting approach.
Since needing to store and compute all the features during feature selection, the feature weighting-based boosting approach is
computationally expensive.
Furthermore, Parag et al. \citeyear{Parag-Porikli-Elgammal-CVPR2008} build a
feature weighting-based BDAM for object tracking, where
the weak classifiers themselves are adaptively
modified to adapt to scene changes.
Namely, the parameters of the weak classifiers are adaptively
changed instead of replacement when the new data arrive.
The common property of the feature weighting-based BDAMs
is that they depend on a fixed number of weak classifiers.
However, this property may restrict the flexibility of the trackers in practice.

\item Dynamic ensemble-based BDAMs.
The conventional BDAMs need to construct a fixed number of
weak learners in advance, and select these weak learners iteratively
as the boosting procedure proceeds. However, due to the time-varying
property of visual object tracking, they are incapable of
effectively adapting to dynamic object appearance changes.
To address this problem, a dynamic ensemble-based BDAM~\cite{Visentini-Snidaro-Foresti-VS2008} is proposed
to dynamically construct and update the set of
weak classifiers according to the ensemble error value.

\item Noise-insensitive BDAMs. To make visual object tracking more robust to noise corruption,
a set of BDAMs are proposed in the literature.
For instance,
Leistner et al.~\citeyear{Leistner-Saffari-Roth-Bischof-OLCV2009}
point out that the convex loss functions typically used in boosting
are highly sensitive to random noise.
To enhance  robustness, Leistner et al.~\citeyear{Leistner-Saffari-Roth-Bischof-OLCV2009}
develop
a generic BDAM called online GradientBoost, which
contains a set of noise insensitive loss functions.
In essence, this BDAM is an extension of the GradientBoost
algorithm~\cite{Friedman-Annal2001} and works similarly to
the AnyBoost algorithm~\cite{Mason-Baxter-Bartlett-Frean-MIT1999}.

\item Particle filtering integration-based BDAMs.
To make visual object tracking more efficient,  researchers embed feature selection
into the particle filtering process.
For example, Wang et al.~\citeyear{Wang-Chen-Gao2005} and Okuma et al.~\citeyear{Okuma-Taleghani-Freitas-Little-Lowe-ECCV2004}
propose two online feature
selection-based BDAMs
using particle filtering, which generate the candidate state set of a tracked object,
and the classification results of AdaBoost is used to determine the final state.

\item Transfer learning-based BDAMs. Typically, most existing BDAMs have an underlying
assumption that the training samples collected from the current frame
follow a similar distribution to those from the last frame.
However, this assumption is often violated when the
``drift'' problem takes place. To address the ``drift'' problem,
a number of novel BDAMs~\cite{wu2012new,luo2011robust} are proposed to categorize the samples into two classes:
auxiliary samples (obtained in the last frames) and target
samples (generated in the current frame). By exploring the intrinsic
proximity relationships among these samples, the proposed BDAMs
are capable of effectively transferring the discriminative
information on auxiliary samples to
the  discriminative learning process using  the current
target samples, leading to robust tracking results.

\end{itemize}

\begin{figure*}[t]
\begin{center}
\includegraphics[width=0.71\linewidth]{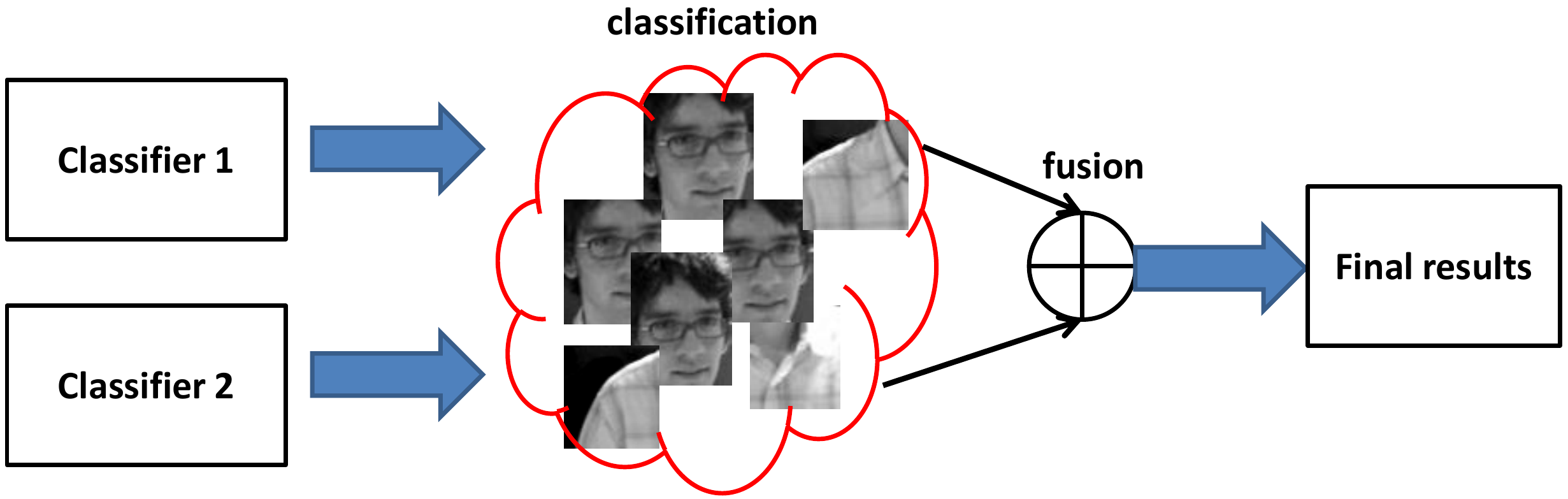}
\end{center}
\vspace{-0.33cm}
   \caption{Illustration of a typical co-learning problem.}
    \label{fig:co-learning}
\end{figure*}

\subsubsection{Co-learning single-instance BDAMs}

In general, the self-learning BDAMs suffer from the ``model drift'' problem
due to their error accumulation caused by using the self-learning strategy.
In order to address this problem,
researchers adopt the semi-supervised learning techniques~\cite{Zhu-SemiL-TR2005} for visual object tracking.
For instance, Grabner et al.~\citeyear{Grabner-Grabner-Bischof-ECCV2008} develop
a BDAM based on semi-supervised online boosting.
Its main idea is to formulate the boosting update process in a semi-supervised
manner as a fused decision of a given prior and an online
classier, as illustrated in Fig.~\ref{fig:co-learning}.
Subsequently, Liu et al.~\citeyear{Liu-Cheng-Lu-ICCV2009}
make use of the co-training strategy to online learn
each weak classifier in boosting instead of only the final
strong classifier. The co-training strategy dynamically generates a series of
unlabeled samples for progressively modifying
the weak classifiers, leading to the robustness to
environmental changes.
It is proven that the co-training strategy can minimize the boosting
error bound in theory.

\subsubsection{{Multi-instance BDAMs}}

\begin{figure*}[t]
\begin{center}
\includegraphics[width=0.7\linewidth]{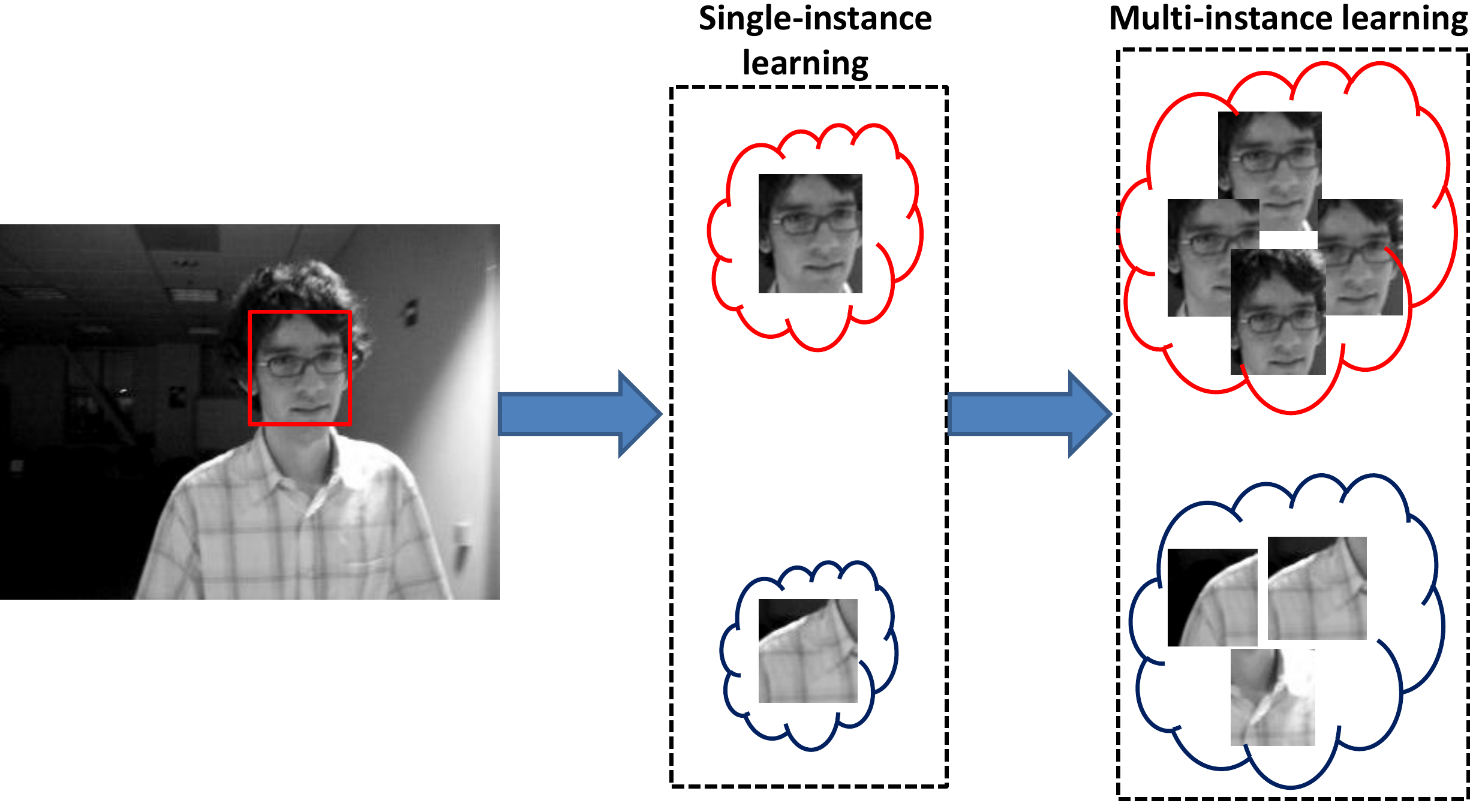}
\end{center}
\vspace{-0.33cm}
   \caption{Illustration of single-instance multi-instance learning. The left part shows the tracking result;
   the middle part displays the positive and negative samples used by the single-instance learning;
   and the right part exhibits the positive and negative sample bags used by the multi-instance learning.}
    \label{fig:multi-instance} \vspace{-0.23cm}
\end{figure*}

To deal with the underlying ambiguity of object localization, multiple instance learning is used for object tracking, as illustrated
in Fig.~\ref{fig:multi-instance}.
In principle, it represents an object by a set of image patches around the tracker location.

\begin{itemize}
\item Self-learning multi-instance BDAMs.
For example, Babenko et al.~\citeyear{Babenko-Yang-Belongie-CVPR2009} represent an object by a set of
image patches, which correspond to an instance bag with each instance being an image patch.
Based on online multiple instance boosting, a tracking system is developed to
characterize the ambiguity of object localization in an online manner.
The tracking system assumes that
all positively labelled instances are truly positive, but this assumption is sometimes violated in practice.
Furthermore, the tracking system trains the weak
classifiers based only on the current frame, and is
likely to be over-fitting.
Instead of equally treating the samples in each bag~\cite{Babenko-Yang-Belongie-CVPR2009}, Zhang et al.~\citeyear{zhang2012real_mil}
propose an online weighted multiple instance tracker, which incorporates the sample importance information
(i.e., the samples closer to the current tracker location are of greater importance)
into the online multi-instance boosting learning process, resulting in robust tracking results.
To characterize the cumulative loss of the weak
classifiers across multiple frames instead of the current frame,
Li et al.~\citeyear{Li-Kwok-Lu-cvpr2010} propose an online
multi-instance BDAM using the strong convex elastic net regularizer instead of
the $\ell_{1}$ regularizer, and further prove that the proposed
multiple instance learning (MIL) algorithm
has a cumulative regret
(evaluating the cumulative loss of the online algorithm) of $\mathcal{O}(\sqrt{T})$ with $T$ being the number of boosting iterations.

\item Co-learning multi-instance BDAMs. Zeisl et al.~\citeyear{Zeisl-Leistner-Saffari-Bischof-CVPR2010} and Li et al.~\citeyear{lissocbt}
combine the advantages of semi-supervised learning and multiple instance
learning in the process of designing a BDAM.
Semi-supervised learning can incorporate more prior
information,
and multiple instance learning focuses on
the uncertainty about where to select positive
samples for model updating.
\end{itemize}

\vspace{-0.15cm}
\subsubsection{{Discussion}}

As mentioned previously, BDAMs can be roughly classified into:
self-learning based and co-learning based ones.
Self-learning based BDAMs adopt the self-learning strategy
to learn object/non-object classifiers. They utilize previously learnt classifiers
to select ``positive'' and ``negative'' training samples,
and then update the current classifiers with the selected training samples.
As a result, tracking errors may be gradually accumulated.  In order to tackle this problem, co-learning based BDAMs
are developed to capture the discriminative information from many
unlabeled samples in each frame. They generally employ semi-supervised co-learning techniques
to update the classifiers with both labeled and unlabeled samples in an interleaved
manner, resulting in more robust tracking results.

On the other hand, conventional BDAMs take a single-instance
strategy for visual representation, i.e., one image patch for each object.
The drawback of this single-instance visual representation is to rely heavily on
exact object localization, without which the tracking performance can be greatly degraded
because of the sub-optimal training sample selection. To address this issue, MIL
is introduced to visual object tracking. It takes into account of the inherent ambiguity of object localization,
representing an object by a set of image patches around the tracker location.
As a result, the MIL-based tracking algorithms can achieve robust tracking results, but may
lose accuracy if the image patches do not precisely capture the object appearance information.

However, all BDAMs need to construct a huge local feature pool for feature selection,
leading to a low computational speed. Additionally,  they usually obtain a local optimal solution to object tracking
because of their focus on local features rather than global features.

\vspace{-0.15cm}
\subsection{{SVM-based discriminative appearance models (SDAMs)}}

SDAMs aim to learn margin-based discriminative SVM classifiers for
maximizing inter-class separability.
SDAMs are able to discover and remember informative samples as
support vectors for object/non-object classification, resulting in a strong discriminative power.
Effective kernel selection and efficient kernel
computation play an importance role in designing robust SDAMs.
According to the used learning mechanisms, SDAMs are typically based on self-learning SDAMs and co-learning
SDAMs.

\begin{itemize}
\item Self-learning SDAMs. In principle, the self-learning SDAMs are to construct SVM classifiers
for object/non-object classification in a self-learning fashion.
For example,
Avidan~\citeyear{Avidan-PAMI2004} proposes an offline SDAM for distinguishing
a target vehicle from a background. Since the SDAM needs substantial prior training data in advance, extending the
algorithm to general object tracking is a difficult task.  Following the work
in~\cite{Avidan-PAMI2004}, Williams et al.~\citeyear{Williams-Blake-Cipolla-PAMI2005} propose a
probabilistic formulation-based SDAM, which allows for propagating
observation distributions over time.  Despite its robustness,  the proposed SDAM needs
to fully encode the appearance variation information, which is impractical in the tracking process.
Tian et al.~\citeyear{Tian-Zhang-Liu-ACCV2007} utilize
an ensemble of linear SVM classifiers to construct a SDAM.
These classifiers can be adaptively weighted according to their discriminative
abilities during different periods, resulting in the robustness to large appearance variations.
The above SDAMs need to heuristically select positive and negative samples surrounding
the current tracker location to update the object/non-object SVM classifier.

\hspace{0.06cm}To avoid the heuristic and unreliable step of training sample selection (usually requiring
accurate estimation of object location), two strategies are adopted in the literature.
One is based on structured output support vector machine (SVM)~\cite{hare2011struck_tracking,yao2012robust}, and the other is
based on ranking SVM~\cite{bai2012robust}. The key idea of these two strategies is to integrate
the structured constraints (e.g., relative ranking or VOC overlap ratio between samples) into the max-margin optimization problem.
For instance, Hare et al.~\citeyear{hare2011struck_tracking}
propose a SDAM based on a kernelized structured SVM,
which involves an infinite number of structured loss (i.e., VOC overlap ratio) based constraints
in the structured output spaces.
In addition, Bai and Tang~\citeyear{bai2012robust}
therefore pose visual object tracking as a weakly supervised ranking problem,
which captures the relative proximity relationships between samples
towards the true target samples.

\item Co-learning SDAMs. In general, the co-learning SDAMs rely on semi-supervised/multi-kernel
learning to construct SVM classifiers for object/non-object classification.
For instance,
Tang et al.~\citeyear{Tang-Brennan-Zhao-ICCV2007}
adopt the co-training SVM technique to
design a semi-supervised tracker.
The disadvantage of this tracker is that it requires several initial frames to generate adequate
labeled samples, resulting in the inflexibility in practice.
Lu et al.~\citeyear{Lu-Zhang-Chen-ACCV2010} and Yang et al.~\citeyear{Yang-Lu-Chen-ACCV2010}
design SVM classifiers using multi-kernel learning (MKL) for visual object tracking.
MKL aims to learn an optimal linear combination of different kernels based on different features, including
the color information and spatial pyramid histogram of visual words.
\end{itemize}

\subsubsection{Discussion}
With the power of max-margin learning, the SDAMs have a good generalization capability of
distinguishing foreground and background, resulting in an effective SVM classifier
for object localization. However, the process of constructing the SDAMs requires a set of
reliable labeled training samples, which is a difficult task due to the influence of some complicated
factors such as
noisy corruptions, occlusions, illumination changes, etc.
Therefore, most existing SDAMs take a heuristic strategy  for
training sample collection (e.g., spatial distance based or classification score based),
which may lead to the instability or even ``drift'' of the tracking process.
To address this issue, the structured SVM is applied to
model the structural relationships (i.e., VOC overlap ratio) between samples,
resulting in a good tracking performance
in terms of generalization and robustness
to noise.
During tracking, a hard assignment of a sample to a class label usually
leads to the classification error accumulation.
To alleviate the issue, the ranking SVM (a weakly supervised learning method) is also introduced into
the tracking process, where the relative ranking
information between samples is incorporated into
the constraints of max-margin learning.

The common point of the above SDAMs is to take a self-learning strategy for object/non-object
classification without considering the discriminative information from
unlabeled data or multiple information sources. Motivated by this, the co-learning SDAMs
are developed to integrate such discriminative information into the
SVM learning process by semi-supervised/multi-kernel
learning.

the co-learning SDAMs emerge

\subsection{Randomized learning-based discriminative appearance models (RLDAMs)}

More recently, randomized learning techniques (e.g., Random
Forest~\cite{Breiman-ML2001,Shotton-Johnson-Cipolla-CVPR2008,Lepetit-Fua-PAMI2006} and
Ferns~\cite{Ozuysal-Calonder-Lepetit-Fua-PAMI2009}) have been successfully introduced into the
vision community.
In principle,
randomized learning techniques can build a diverse classifier ensemble  by performing random input selection
and random feature selection. In contrast to boosting and SVM, they are more computationally efficient,
and easier to be extended for handling multi-class learning problems.
In particular, they can be parallelized so that
multi-core and GPU implementations (e.g., \cite{Sharp-ECCV2008}) can be performed
to greatly reduce the run time.
However, their tracking performance is unstable for different scenes because of their random feature selection.

Inspired by randomized learning, a variety of RLDAMs are proposed in the field of
visual object tracking, including online random forests~\cite{Saffari-Leistner-Santner-Godec-Bischof-OLCV2009,Santner-Leistner-Saffari-Pock-Bischof-cvpr2010},
random naive Bayes classifiers~\cite{Godec-Leistner-Saffari-Bischof-ICPR2010}, and MIForests~\cite{Leistner-Saffari-Bischof-ECCV2010}.
For instance,
Godec et al.~\citeyear{Godec-Leistner-Saffari-Bischof-ICPR2010}
develop a visual object tracking algorithm based on online random naive Bayes classifiers.
Due to the low computational and
memory costs of Random Naive Bayes classifiers,
the developed tracking algorithm
has a powerful real-time capability for
processing long-duration video sequences.
In contrast to online Random Forests~\cite{Saffari-Leistner-Santner-Godec-Bischof-OLCV2009},
the random Naive Bayes classifiers have
a higher computational efficiency and faster convergence in the training phase.
Moreover, Leistner et al.~\citeyear{Leistner-Saffari-Bischof-ECCV2010} present a
RLDAM named MIForests, which uses
multiple instance learning to construct randomized trees and represents
the hidden class labels inside target bags as random variables.

\subsection{Discriminant analysis-based discriminative appearance models (DADAMs)}

Discriminant analysis is a powerful tool for supervised subspace learning.
In principle, its goal is to find a low-dimensional subspace
with a high inter-class separability.
According to the
learning schemes used, it can be split into two branches: conventional discriminant analysis and graph-driven
discriminant analysis. In general, conventional DADAMs are
formulated in a vector space while graph-driven
DADAMs utilize graphs for supervised subspace learning.

\subsubsection{Conventional DADAMs}
Typically, conventional discriminant analysis techniques
can be divided into one of the following two main branches.

\begin{itemize}
\item Uni-modal DADAMs. In principle, uni-modal DADAMs
have a potential assumption that the data for the object class
follow a uni-modal Gaussian distribution.
For instance,
Lin et al.~\citeyear{Lin-Yang-Levinson-ICPR2004} build a DADAM based on incremental Fisher linear discriminant analysis (IFLDA).
This DADAM models the object class as a single Gaussian
distribution, and models the background class as a mixture of Gaussian
distributions.
In~\cite{Nguyen-Smeulders-IJCV2006}, linear discriminant analysis (LDA) is used for discriminative learning
in the local texture feature space obtained by Gabor filtering.
However, there is a potential assumption that the distributions of the object and the background classes
are approximately Gaussian ones with an equal covariance.
Li et al.~\citeyear{Li-Liang-Huang-Jiang-Gao-ICIP2008} construct a DADAM using
the incremental 2DLDA on the 2D image matrices.
Since matrix operations are directly made on these 2D matrices,
the DADAM is computationally efficient.
Moreover, another way of constructing uni-modal DADAMs
is by discriminant metric learning, which aims to
linearly map the original feature space
to a new metric space by a linear projection~\cite{wang2010discriminative,jiang2011tracking,jiang2012order}.
After discriminant metric learning, the similarity
between intra-class samples are minimized while the distance
between inter-class samples are maximized, resulting
in an effective similarity measure for robust object tracking.
Note that the above DADAMs are incapable
of dealing well with the object and background classes
having multi-modal distributions.

\item Multi-modal DADAMs. In essence, multi-modal DADAMs
model the object class and the background class as
a mixture of Gaussian distributions.
For example,
Xu et al.~\citeyear{Xu-Shi-Xu-2008} take advantage of adaptive subclass
discriminant analysis (SDA) (i.e., an extension
to the basic SDA~\cite{Zhu-Martinez-PAMI2006}) for object tracking.
The adaptive SDA first partitions data samples
into several subclasses by a nearest neighbor clustering,
and then runs the traditional LDA for each subclass.
\end{itemize}

\begin{figure*}[t]
\begin{center}
\includegraphics[width=0.7\linewidth]{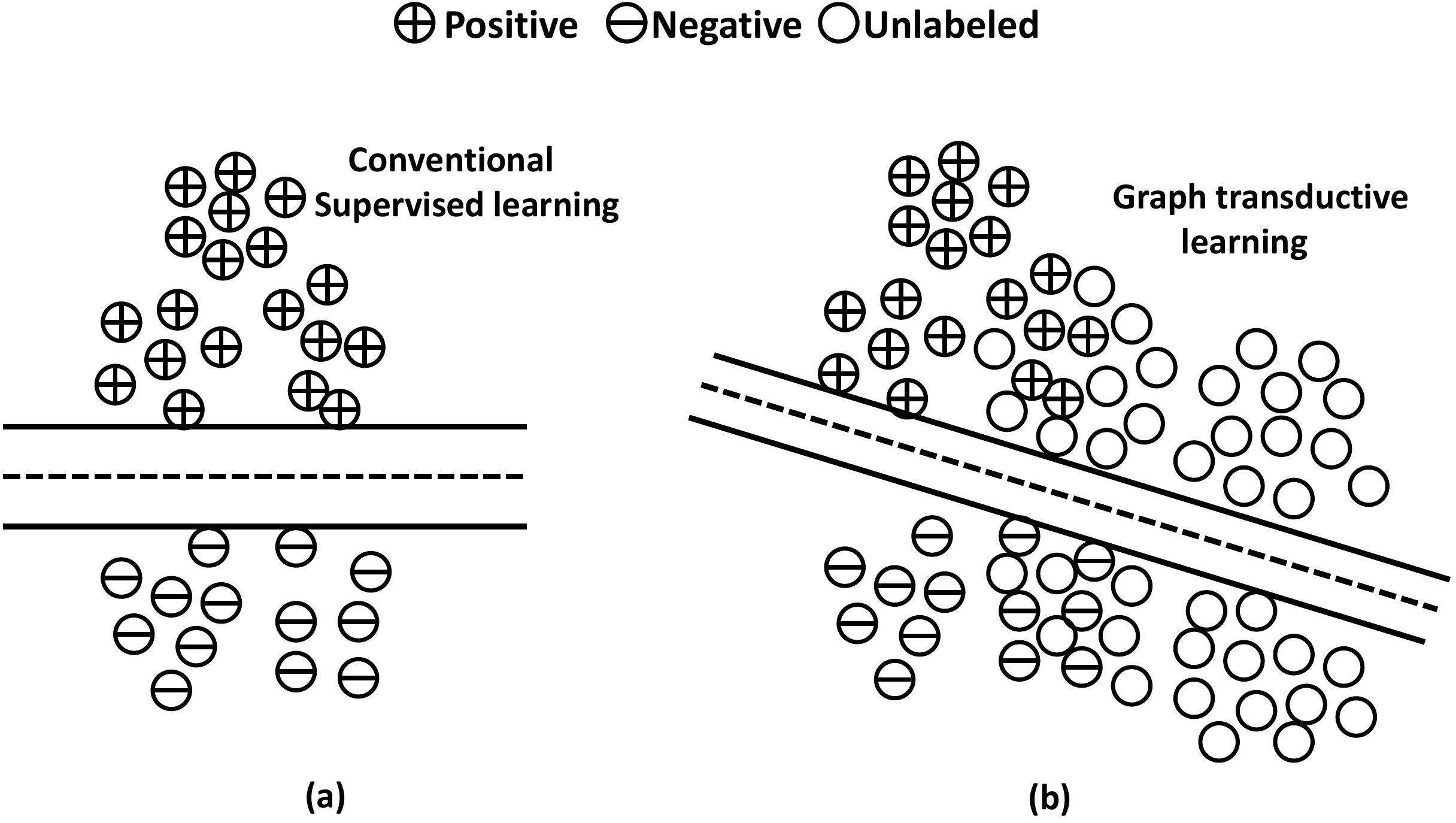}
\end{center}
\vspace{-0.33cm}
   \caption{Illustration of transductive learning. (a) shows the decision hyperplane obtained by the
   conventional supervised learning; and (b) displays the decision hyperplane (further adjusted
   by the unlabeled samples) of transductive learning.}
    \label{fig:transductive_learning}
\end{figure*}

\subsubsection{Graph-driven DADAMs}
Researchers utilize the generalized
graph-based discriminative learning (i.e., graph embedding and graph transductive learning)
to construct a set of DADAMs
for visual object tracking.  Typically, these DADAMs mainly have the following two branches:

\begin{itemize}
\item Graph embedding based DADAMs. In principle, the goal of graph embedding based DADAMs
is to set up a graph-based discriminative model, which
utilizes the graph-based techniques to embed the high-dimensional samples
into a discriminative low-dimensional space
for the object/non-object classification.
For instance,
Zhang et al.~\citeyear{Zhang-Hu-Maybank-Li-ICCV2007}
design a DADAM based on graph embedding-based LDA, which
makes a basic assumption
that the background class is irregularly distributed
with multiple modalities while the object class follows
a single Gaussian distribution.
However, this basic assumption does not hold true
in the case of complex intrinsic and extrinsic object appearance changes.

\item Graph transductive learning based DADAMs. In general, graph transductive learning based DADAMs
aim to utilize the power of graph-based semi-supervised transductive learning
for the likelihood evaluation of the candidate samples belonging to the object class.
They make use of the intrinsic topological information
between the labeled and unlabeled samples to discover
an appropriate decision hyperplane for object/non-object classification, as shown in  Fig.~\ref{fig:transductive_learning}.
For instance,
Zha et al.~\citeyear{Zha-Yang-Bi-PR2010} develop a tracker based on
graph-based
transductive learning.
The tracker utilizes the labeled samples to maximize the inter-class
separability, and the unlabeled samples to
capture the underlying geometric structure of the samples.

\end{itemize}

\subsubsection{Discussion}
The goal of DADAMs is to learn a decision hyperplane to separate
the object class from the background class. However, the traditional DADAMs
perform poorly when both the object class and the background class
have multi-modal statistical distributions. To overcome this limitation, multi-modal
discriminant analysis is adopted to explore the training data distributions by data clustering.
To make a non-linear extension to the conventional DADAMs, graph-based DADAMs are proposed.
These DADAMs try to formulate the problem of discriminant analysis as
that of graph learning such as graph embedding and graph transductive learning.
However, a drawback
is that these algorithms need to retain a large amount of labeled/unlabeled samples for graph learning,
leading to their impracticality for real tracking applications.

\begin{table*}[t]
\vspace{-0.3cm}
\caption{{Summary of representative tracking-by-detection methods using hybrid generative-discriminative learning techniques \vspace{-0.0cm}}}
\label{tab:hybrid_tracking}
\hspace{-0.00cm}
\scalebox{0.68}{
\begin{tabular}{c|c|c|c|c}\hline
Item No.&
\makebox[8.2em]{
\begin{tabular}{c}
References\\
\end{tabular}}
&
\makebox[5.2em]{
\begin{tabular}{c}
Single-layer\\
combination
\end{tabular}}
&

\begin{tabular}{c}
Multi-layer\\
combination
\end{tabular}
&
\begin{tabular}{c}
Used learning\\
techniques
\end{tabular}
 \\\hline\hline
1&
\begin{tabular}{c}
\cite{Kelm-Pal-McCallum-ICPR2006}\\
\end{tabular}
&\Large{$\surd$}
& \Large{$\times$} &
\begin{tabular}{c}
Multi-conditional
learning
\end{tabular}
\\ [0 ex] \hline

2&
\begin{tabular}{c}
\cite{Lin-Ross-Lim-Yang-CVPR2004}\\
\end{tabular}
&\Large{$\surd$}
& \Large{$\times$} &
\begin{tabular}{c}
Combination of PCA \\
and Fisher
LDA
\end{tabular}
\\ [0 ex] \hline

3&
\begin{tabular}{c}
\cite{Grabner-Roth-Bischof-CVPR2007}\\
\end{tabular}
&\Large{$\surd$}
 & \Large{$\times$} &
\begin{tabular}{c}
Combination of boosting\\
and robust PCA
\end{tabular}
\\ [0 ex] \hline

4&
\begin{tabular}{c}
\cite{Yang-Fan-Fan-Wu-TIP2009}\\
\end{tabular}
&\Large{$\surd$}
& \Large{$\times$} &
\begin{tabular}{c}
Discriminative subspace learning\\
using positive and negative data
\end{tabular}
\\ [0 ex] \hline

5&
\begin{tabular}{c}
\cite{Everingham-Zisserman-ICCV2005}\\
\end{tabular}
&\Large{$\times$}
& \Large{$\surd$} &
\begin{tabular}{c}
Combination of a tree-structured
classifier \\
and a Lambertian lighting model
\end{tabular}
\\ [0 ex] \hline

6&
\begin{tabular}{c}
\cite{Shen-Kim-Wang-TCSVT2010}\\
\end{tabular}
&\Large{$\times$}
& \Large{$\surd$} &
\begin{tabular}{c}
Combination of SVM learning\\
and kernel density estimation
\end{tabular}
\\ [0 ex] \hline

7&
\begin{tabular}{c}
\cite{Lei-Ding-Wang-SMCB2008}\\
\end{tabular}
&\Large{$\times$}
& \Large{$\surd$} &
\begin{tabular}{c}
Three-layer combination of \\
relevance vector machine
and GMM:\\
learner combination (Layer 1)\\
classifier combination (Layer 2)\\
decision combination (Layer 3)\\
\end{tabular}
\\ [0 ex] \hline

8&
\begin{tabular}{c}
\cite{Yu-Dinh-Medioni-ECCV2008}\\
\end{tabular}
&\Large{$\times$}
& \Large{$\surd$} &
\begin{tabular}{c}
Combination of the  constellation\\
model
and  fisher kernels\\
\end{tabular}
\\ [0 ex] \hline

\end{tabular}
}
\end{table*}

\subsection{{Codebook learning-based discriminative appearance models (CLDAMs)}}

In principle, CLDAMs need to construct the foreground and background codebooks
to adaptively capture the dynamic appearance information from the foreground and background.
Recently,
Yang et al.~\citeyear{Yang-Lu-Chen-ICPR2010} construct two
codebooks of image patches using two different features: RGB and LBP features, leading to the
robustness in handling occlusion, scaling, and rotation.  To capture more discriminative information,
an adaptive class-specific codebook~\cite{Gall-Razavi-Gool-BMVC2010} is built
for instance tracking. The codebook encodes the information on spatial distribution and
appearance of object parts, and
can be converted to a more
instance-specific codebook in a probabilistic way (i.e., probabilistic votes for the object
instance).  Inspired by the
tracking-by-detection idea, Andriluka et al.~\citeyear{Andriluka-Roth-Schiele-CVPR2008}
establish object-specific codebooks, which are constructed by clustering
local features (i.e., shape context feature descriptors and Hessian-Laplace interest points)
extracted from a set of training images. These codebooks are then embedded into a part-based model for
pedestrian detection.

Therefore, CLDAMs often consider the discriminative
information not only from the background but also from other object instances.
However, it is very difficult to construct a universal codebook for different scenes or objects.  As
a result, it is necessary to collect different training samples for different scenes or objects,
leading to inflexibility in practice. In addition, determining the codebook size is a
difficult task in practice.

\subsection{Hybrid generative-discriminative appearance models (HGDAMs)}

As discussed in~\cite{Ulusoy-Bishop-CVPR2005}, the generative and the discriminative models have their own advantages
and disadvantages, and are complementary to each other to some extent.
Consequently, much effort has been made to propose a variety of
hybrid generative-discriminative models for combining
the benefits of both the generative and the discriminative models in visual object tracking.
These hybrid generative-discriminative models aim to combine the generative and the discriminative
models in a single-layer or multi-layer  manner.

\subsubsection{HGDAMs via single-layer combination}

HGDAMs via single-layer combination aim to fuse the generative and the discriminative
models at the same layer. They attempt to fuse the confidence scores
of the generative and the discriminative models
to generate better tracking results than just using them individually.
Typically, they have two kinds of combination mechanisms:
decision-level combination and intermediate-level combination.

\begin{itemize}
\item HGDAMs via decision-level combination. In principle, such HGDAMs focus on how to effectively
fuse the confidence scores from the generative and the discriminative models.
For instance,  a linear fusion strategy~\cite{Kelm-Pal-McCallum-ICPR2006}
is taken to combine
the log-likelihood of discriminative and
generative models for pixel classification.
It is
pointed out in~\cite{Kelm-Pal-McCallum-ICPR2006}  that the performance of the combined
generative-discriminative models is associated with
a balance between the purely
generative and purely discriminative ones.
In addition,
Lin et al. \cite{Lin-Ross-Lim-Yang-CVPR2004}
propose a HGDAM that is a
generalized version of the Fisher Linear Discriminant Analysis.  This
HGDAM consists of two components: the observation sub-model and the
discriminative sub-model.

\item HGDAMs via intermediate-level combination. In principle, the HGDAMs via intermediate-level combination aim to simultaneously
utilize both low-level features and high-level confidence scores
from the generative and the discriminative models.
For instance, Yang et al.~\citeyear{Yang-Fan-Fan-Wu-TIP2009}
impose three data-driven constraints on the proposed object appearance model:
(1) negative data;
(2) bottom-up pair-wise data constraints; and (3) adaptation dynamics.
As a result,
the object appearance model can greatly ameliorate the problem of
adaptation drift and can achieve good tracking performances in various
non-stationary scenes. Furthermore, Grabner et al.~\citeyear{Grabner-Roth-Bischof-CVPR2007} propose a
HGDAM based on a boosting algorithm called Eigenboosting, which requires visual
features to be discriminative with reconstructive abilities at the same time.  In principle,
eigenboosting aims to minimize a modified boosting error-function in which the generative
information (i.e., eigenimages generated from Haar-like binary basis-functions using robust PCA)
is integrated as a multiplicative prior.
\end{itemize}

\begin{table*}[t]
\vspace{-0.3cm}
\caption{{Summary of the publicly available tracking resources}}
\label{tab:tracking_datasets}
\hspace{-0.16cm}
\scalebox{0.56}{
\begin{tabular}{c|*{6}{c}}\hline
Item No. &Name & Dataset  & Ground truth & Source code   & Web link \\\hline\hline
1&
\begin{tabular}{c}
Head track\\
\cite{Birchfield-CVPR1998}
\end{tabular}
 & \large{$\surd$} & \large{$\times$} & \large{$\surd$} &
www.ces.clemson.edu/$\sim$stb/research/headtracker/seq/\\ \hline  %

2&
\begin{tabular}{c}
Fragment tracker\\
\cite{Adam-Rivlin-Shimshoni-cvpr2006}
\end{tabular}
& \large{$\surd$} & \large{$\surd$} & \large{$\surd$} & www.cs.technion.ac.il/$\sim$amita/fragtrack/fragtrack.htm \\ \hline

3&
\begin{tabular}{c}
Adaptive tracker\\
\cite{Jepson-Fleet-Yacoob-PAMI2003}
\end{tabular}
& \large{$\surd$} & \large{$\times$} & \large{$\times$} & www.cs.toronto.edu/vis/projects/adaptiveAppearance.html \\ \hline

4&
\begin{tabular}{c}
PCA tracker\\
\cite{Limy-Ross17}
\end{tabular}
& \large{$\surd$} & \large{$\surd$} & \large{$\surd$} & www.cs.utoronto.ca/$\sim$dross/ivt/\\ \hline

5&
\begin{tabular}{c}
KPCA tracker\\
\cite{Chin-Suter-TIP2007}
\end{tabular}
& \large{$\times$} & \large{$\times$} & \large{$\surd$} &  cs.adelaide.edu.au/$\sim$tjchin/ \\ \hline

6&
\begin{tabular}{c}
$\ell_{1}$ tracker\\
\cite{Meo-Ling-ICCV09}
\end{tabular}
& \large{$\times$} & \large{$\times$} & \large{$\surd$} & www.ist.temple.edu/$\sim$hbling/code$\_$data.htm \\ \hline

7&
\begin{tabular}{c}
Kernel-based tracker\\
\cite{Shen-Kim-Wang-TCSVT2010}
\end{tabular}
& \large{$\times$} & \large{$\times$} & \large{$\surd$} & code.google.com/p/detect/ \\ \hline

8&
\begin{tabular}{c}
Boosting tracker\\
\cite{Grabner-Bischof-CVPR2006}
\end{tabular}
& \large{$\surd$} & \large{$\times$} & \large{$\surd$} & www.vision.ee.ethz.ch/boostingTrackers/ \\ \hline

9&
\begin{tabular}{c}
MIL tracker\\
\cite{Babenko-Yang-Belongie-CVPR2009}
\end{tabular}
& \large{$\surd$} & \large{$\surd$} & \large{$\surd$} & vision.ucsd.edu/$\sim$bbabenko/project$\_$miltrack.shtml \\ \hline

10&
\begin{tabular}{c}
MIForests tracker\\
\cite{Leistner-Saffari-Bischof-ECCV2010}
\end{tabular}
 & \large{$\surd$} & \large{$\surd$} & \large{$\surd$} & www.ymer.org/amir/software/milforests/ \\ \hline

11&
\begin{tabular}{c}
Boosting+ICA tracker\\
\cite{Yang-Lu-Chen-ICIP2010}
\end{tabular}
& \large{$\times$} & \large{$\times$} & \large{$\surd$} & ice.dlut.edu.cn/lu/publications.html \\ \hline

12&
\begin{tabular}{c}
Appearance-adaptive tracker\\
\cite{Zhou-Chellappa-Moghaddam6}
\end{tabular}
& \large{$\times$} & \large{$\times$} & \large{$\surd$} & www.umiacs.umd.edu/$\sim$shaohua/sourcecodes.html \\ \hline

13&
\begin{tabular}{c}
Tracking with histograms \\
and articulating blocks\\
\cite{Nejhum-Ho-Yang-CVIU2010}
\end{tabular}
& \large{$\surd$} & \large{$\surd$} & \large{$\surd$} & www.cise.ufl.edu/$\sim$smshahed/tracking.htm \\ \hline

14&
\begin{tabular}{c}
Visual tracking decomposition\\
\cite{Kwon-Lee-CVPR2010}
\end{tabular}
& \large{$\surd$} & \large{$\surd$} & \large{$\surd$} & cv.snu.ac.kr/research/$\sim$vtd/\\ \hline

15&
\begin{tabular}{c}
Structural SVM  tracker\\
\cite{hare2011struck_tracking}
\end{tabular}
& \large{$\times$} & \large{$\times$} & \large{$\surd$} & www.samhare.net/research/struck\\ \hline %

16&
\begin{tabular}{c}
PROST tracker\\
\cite{Santner-Leistner-Saffari-Pock-Bischof-cvpr2010}
\end{tabular}
& \large{$\surd$} & \large{$\surd$} & \large{$\surd$} & gpu4vision.icg.tugraz.at/index.php?content=subsites/prost/prost.php\\ \hline

17&
\begin{tabular}{c}
Superpixel tracker\\
\cite{Superpixeltrackingiccv2011}
\end{tabular}
& \large{$\surd$} & \large{$\surd$} & \large{$\surd$} & faculty.ucmerced.edu/mhyang/papers/iccv11a.html\\ \hline

18&
\begin{tabular}{c}
KLT feature tracker\\
\cite{Lucas-Kanade-IJCAI1981}
\end{tabular}
&  \large{$\times$} & \large{$\times$} & \large{$\surd$} & www.ces.clemson.edu/$\sim$stb/klt/ \\ \hline

19&
\begin{tabular}{c}
Deformable contour tracker\\
\cite{Vaswani-Rathi-Yezzi-Tannenbaum-TIP2008}
\end{tabular}
 & \large{$\surd$} & \large{$\times$} & \large{$\surd$} & home.engineering.iastate.edu/$\sim$namrata/research/ContourTrack.html$\#$code \\ \hline %

20&
\begin{tabular}{c}
Condensation tracker\\
\cite{Isard-Blake-IJCV1998}
\end{tabular}
& \large{$\surd$} & \large{$\times$} & \large{$\surd$} & www.robots.ox.ac.uk/$\sim$misard/condensation.html \\ \hline %

21&
\begin{tabular}{c}
Motion tracking\\
\cite{Stauffer-Grimson-PAMI2000}
\end{tabular}
& \large{$\surd$} & \large{$\times$} & \large{$\surd$} & www.cs.berkeley.edu/$\sim$flw/tracker/ \\ \hline %

22&
\begin{tabular}{c}
Mean shift tracker
\end{tabular}
 & \large{$\times$}
 & \large{$\times$} & \large{$\surd$} & www.cs.bilkent.edu.tr/$\sim$ismaila/MUSCLE/MSTracker.htm \\ \hline

23&
\begin{tabular}{c}
Tracking-Learning-Detection Tracker
\end{tabular}
 & \large{$\surd$}
 & \large{$\times$} & \large{$\surd$} & info.ee.surrey.ac.uk/Personal/Z.Kalal/tld.html \\ \hline

24&
CAVIAR sequences
& \large{$\surd$} & \large{$\surd$} & \large{$\times$} & homepages.inf.ed.ac.uk/rbf/CAVIARDATA1/\\ \hline

25&
PETS sequences & \large{$\surd$} & \large{$\surd$} & \large{$\times$} &
\begin{tabular}{c}
www.hitech-projects.com/euprojects/cantata/datasets$\_$cantata/dataset.html
\end{tabular}
\\ \hline

26&
SURF & \large{$\times$}
 & \large{$\times$} & \large{$\surd$} & people.ee.ethz.ch/$\sim$surf/download$\_$ac.html \\ \hline

27&
XVision visual tracking &  \large{$\times$} & \large{$\times$} & \large{$\surd$} & peipa.essex.ac.uk/info/software.html \\ \hline

28&
The Machine Perception Toolbox &  \large{$\times$} & \large{$\times$} & \large{$\surd$} & mplab.ucsd.edu/grants/project1/free-software/MPTWebSite/introduction.html \\ \hline

29&
Compressive Tracker~\cite{zhangCompressiveTracking2012} & \large{$\surd$} & \large{$\surd$} & \large{$\surd$} &  www4.comp.polyu.edu.hk/$\sim$cslzhang/CT/CT.htm \\ \hline

30&
\begin{tabular}{c}
Structural local sparse tracker\\
\cite{jia2012visual}
\end{tabular}
& \large{$\surd$} & \large{$\surd$} & \large{$\surd$} &  ice.dlut.edu.cn/lu/Project/cvpr12\_jia\_project/cvpr12\_jia\_project.htm \\ \hline

31&
\begin{tabular}{c}
Sparsity-based collaborative tracker\\
\cite{zhong2012robust}
\end{tabular}
& \large{$\surd$} & \large{$\surd$} & \large{$\surd$} &  ice.dlut.edu.cn/lu/Project/cvpr12\_scm/cvpr12\_scm.htm \\ \hline

32&
\begin{tabular}{c}
Multi-task sparse tracker\\
\cite{zhang2012robust-multitask}
\end{tabular}
& \large{$\times$} & \large{$\times$} & \large{$\surd$} &  sites.google.com/site/zhangtianzhu2012/publications \\ \hline

33&
\begin{tabular}{c}
APG $\ell_{1}$ tracker
\cite{bao2012real}
\end{tabular}
& \large{$\surd$} & \large{$\surd$} & \large{$\surd$} &  www.dabi.temple.edu/$\sim$hbling/code\_data.htm\#L1\_Tracker \\ \hline

34&
\begin{tabular}{c}
Structured keypoint tracker\\
\cite{hare2012efficient}
\end{tabular}
& \large{$\surd$} & \large{$\surd$} & \large{$\surd$} &  www.samhare.net/research/keypoints \\ \hline

35&
\begin{tabular}{c}
Spatial-weighted MIL tracker\\
\cite{zhang2012real_mil}
\end{tabular}
& \large{$\times$} & \large{$\times$} & \large{$\surd$} &  code.google.com/p/online-weighted-miltracker/ \\ \hline

\end{tabular}
}
\end{table*}

\subsubsection{{HGDAMs via multi-layer combination}}

In principle, the goal of the HGDAMs via multi-layer combination is to
combine the information from the generative and discriminative models
at multiple layers. In general, such HGDAMs can be divided
into two classes: HGDAMs via sequential combination and
HGDAMs via interleaving combination.

\begin{itemize}
\item HGDAMs via sequential combination. In principle, the HGDAMs via sequential combination
aim to fuse the benefits of the generative
and discriminative
models in a sequential manner.
Namely, they use the decision output of one model
as the input of the other model. For example,
Everingham and Zisserman~\cite{Everingham-Zisserman-ICCV2005}
combine
generative and discriminative head models.
A discriminative tree-structured classifier
is trained to make efficient detection and pose estimation
over a large pose space with three degrees of freedom.
Subsequently, a generative head model is used for the identity
verification.
Moreover,
Shen et al.~\citeyear{Shen-Kim-Wang-TCSVT2010} develop a generalized
kernel-based HGDAM which learns
a dynamic visual representation by online SVM learning.
Subsequently, the learned visual representation
is incorporated into the standard MS tracking procedure.
Furthermore,
Lei et al.~\citeyear{Lei-Ding-Wang-SMCB2008}
propose a HGDAM using sequential Bayesian learning.
The proposed tracking algorithm consists of three modules.
In the first module, a fast relevance vector machine algorithm
is used to learn a discriminative classifier.
In the second module, a sequential Gaussian mixture model is
learned for visual representation.
In the third module, a model combination
mechanism with a three-level hierarchy is discussed, including
the learner combination (at level one),
classifier combination (at level two), and decision combination
(at level three).

\item HGDAMs via interleaving combination. In principle, the goal of the HGDAMs via interleaving combination
is to combine the discriminative-generative information in a multi-layer interleaving manner.
Namely, the decision output of one model is used to guide the learning task of the other model and vice versa.
For instance, Yu et al.~\citeyear{Yu-Dinh-Medioni-ECCV2008} utilize a co-training strategy to
combine the information from a SVM classifier and a generative multi-subspace model~\cite{Lee-Kriegman14}
in a multi-layer interleaving manner.
\end{itemize}

\begin{table*}[t]
\vspace{-0.3cm}
\caption{{Qualitative comparison of visual representations (Symbols $\surd$ and $\times$ mean that
the visual representation can or cannot cope with the situations of occlusions, illumination changes, and shape deformations, respectively.)}}
\label{tab:feature_evaluations}
\hspace{0.0cm}
\scalebox{0.59}{
\begin{tabular}{ c|c|c|c|c|c|c }\hline
Item No.&
\begin{tabular}{c}
Visual representations
\end{tabular}
&
\begin{tabular}{c}
What to track
\end{tabular}
&
\begin{tabular}{c}
Speed
\end{tabular}
&
\begin{tabular}{c}
Occlusion
\end{tabular}
&
\begin{tabular}{c}
Illumination
\end{tabular}
&
\begin{tabular}{c}
Shape deformation
\end{tabular}
 \\\hline\hline

1&
\begin{tabular}{c}
Vector-based raw pixel representation\\
\cite{Limy-Ross17}\\
\end{tabular}
&
\begin{tabular}{c}
rectangle
\end{tabular}
&
high
&
$\times$
&
$\times$
&
$\times$
\\ [0 ex] \hline

2 &
\begin{tabular}{c}
Matrix-based raw pixel representation\\
\cite{lixi-iccv2007}
\end{tabular}
&
\begin{tabular}{c}
rectangle
\end{tabular}
&
high
&
$\times$
&
$\times$
&
$\times$
\\ \hline

3&
\begin{tabular}{c}
Multi-cue raw pixel representation\\
(i.e., color, position, edge)
\cite{Wang-Suter-Schindler-PAMI2007}\\
\end{tabular}
&
\begin{tabular}{c}
rectangle
\end{tabular}
&
\begin{tabular}{c}
moderate
\end{tabular}
&
$\surd$
&
$\times$
&
$\times$
\\ \hline

4&
\begin{tabular}{c}
Multi-cue spatial-color histogram representation \\
(i.e., joint histogram in (x, y, R, G, B))\\
\cite{Georgescu-Meer-PAMI2004}\\
\end{tabular}
&
\begin{tabular}{c}
rectangle
\end{tabular}
&
high
&
$\times$
&
$\times$
&
$\surd$
\\ \hline

5&
\begin{tabular}{c}
Multi-cue spatial-color histogram representation \\
(i.e., patch-division histogram)\\
\cite{Adam-Rivlin-Shimshoni-cvpr2006}
\end{tabular}
&
\begin{tabular}{c}
rectangle
\end{tabular}
&
high
&
$\surd$
&
$\times$
&
$\surd$
\\ \hline

6&
\begin{tabular}{c}
covariance representation\\
\cite{Porikli-Tuzel-Meer-cvpr2006,lixi-cvpr2008}\\
\cite{hu2012single,wu2012realtip}
\end{tabular}
&
\begin{tabular}{c}
rectangle
\end{tabular}
&
moderate
&
$\times$
&
$\surd$
&
$\surd$
\\ \hline

7&
\begin{tabular}{c}
Wavelet filtering-based representation\\
\cite{Li-Zhang-Huang-Tan-ICIP2009}
\end{tabular}
&
\begin{tabular}{c}
rectangle
\end{tabular}
&
slow
&
$\surd$
&
$\surd$
&
$\surd$
\\ \hline

8&
\begin{tabular}{c}
\cite{Cremers-PAMI2006,sun2011novel}\\
Active contour representation
\end{tabular}
&
\begin{tabular}{c}
contour
\end{tabular}
&
slow
&
$\surd$
&
$\times$
&
$\surd$
\\ \hline

9&
\begin{tabular}{c}
Local feature-based represnetation\\
(local templates)\\
\cite{lin2007hierarchical}
\end{tabular}
&
\begin{tabular}{c}
rectangle
\end{tabular}
&
moderate
&
$\surd$
&
$\surd$
&
$\surd$
\\ \hline

10&
\begin{tabular}{c}
Local feature-based represnetation\\
(MSER features)\\
\cite{Tran-Davis-iccv2007}\\
\end{tabular}
&
\begin{tabular}{c}
irregular\\
regions
\end{tabular}
&
slow
&
$\surd$
&
$\times$
&
$\surd$
\\ \hline

11&
\begin{tabular}{c}
\cite{Zhou-Yuan-Shi-CVIU2009}\\
Local feature-based represnetation\\
(SIFT features)
\end{tabular}
&
\begin{tabular}{c}
interest\\
points
\end{tabular}
&
slow
&
$\surd$
&
$\surd$
&
$\surd$
\\ \hline

12&
\begin{tabular}{c}
Local feature-based represnetation\\
(SURF features)\\
\cite{He-Yamashita-Lu-Lao-ICCV2009}
\end{tabular}
&
\begin{tabular}{c}
interest\\
points
\end{tabular}
&
moderate
&
$\surd$
&
$\surd$
&
$\surd$
\\ \hline

13&
\begin{tabular}{c}
Local feature-based represnetation\\
(Corner features)\\
\cite{Kim-CVPR2008}\\
\end{tabular}
&
\begin{tabular}{c}
interest\\
points
\end{tabular}
&
moderate
&
$\surd$
&
$\surd$
&
$\surd$
\\ \hline

14&
\begin{tabular}{c}
Local feature-based represnetation\\
(Saliency detection-based features)\\
\cite{Fan-Wu-Dai-ECCV2010}\\
\end{tabular}
&
\begin{tabular}{c}
saliency\\
patches
\end{tabular}
&
slow
&
$\surd$
&
$\surd$
&
$\surd$
\\ \hline

\end{tabular}
}
\end{table*}

\begin{table*}[t]
\vspace{-0.3cm}
\caption{{Qualitative comparison of representative statistical modeling based appearance models.}}
\label{tab:model_evaluations}
\hspace{-0.1cm}
\scalebox{0.56}{
\begin{tabular}{ c|c|c|c|c|c|c }\hline
Item No.&
\begin{tabular}{c}
Statistical modeling-based\\
appearance models
\end{tabular}
&
\begin{tabular}{c}
Domain
\end{tabular}
&
\begin{tabular}{c}
Speed
\end{tabular}
&
\begin{tabular}{c}
Memory usage
\end{tabular}
&
\begin{tabular}{c}
Online adaptability
\end{tabular}
&
\begin{tabular}{c}
Discriminability
\end{tabular}
 \\\hline\hline

1&
\begin{tabular}{c}
Linear subspace models
\end{tabular}
&
\begin{tabular}{c}
manifold\\
learning
\end{tabular}
&
\begin{tabular}{c}
fast
\end{tabular}
&
low
&
strong
&
weak
\\ [0 ex] \hline

2&
\begin{tabular}{c}
Nonlinear subspace models
\end{tabular}
&
\begin{tabular}{c}
manifold\\
learning
\end{tabular}
&
\begin{tabular}{c}
slow
\end{tabular}
&
high
&
weak
&
moderate
\\ [0 ex] \hline

3&
\begin{tabular}{c}
Mixture models
\end{tabular}
&
\begin{tabular}{c}
Parametric\\
density estimation
\end{tabular}
&
\begin{tabular}{c}
moderate
\end{tabular}
&
low
&
strong
&
moderate
\\ [0 ex] \hline

4&
\begin{tabular}{c}
Kernel-based models
\end{tabular}
&
\begin{tabular}{c}
Nonparametric\\
density estimation
\end{tabular}
&
\begin{tabular}{c}
fast
\end{tabular}
&
low
&
weak
&
weak
\\ [0 ex] \hline

5&
\begin{tabular}{c}
Boosting-based\\
appearance models
\end{tabular}
&
\begin{tabular}{c}
ensemble learning
\end{tabular}
&
\begin{tabular}{c}
moderate
\end{tabular}
&
low
&
strong
&
strong
\\ [0 ex] \hline

6&
\begin{tabular}{c}
SVM-based\\
appearance models
\end{tabular}
&
\begin{tabular}{c}
Maximum margin\\
learning
\end{tabular}
&
\begin{tabular}{c}
slow
\end{tabular}
&
high
&
strong
&
strong
\\ [0 ex] \hline

7&
\begin{tabular}{c}
Randomized learning\\
based appearance models
\end{tabular}
&
\begin{tabular}{c}
classifier ensemble\\
based on random input selection\\
and random feature selection
\end{tabular}
&
\begin{tabular}{c}
fast
\end{tabular}
&
high
&
strong
&
weak
\\ [0 ex] \hline

8&
\begin{tabular}{c}
Discriminant analysis\\
based appearance models
\end{tabular}
&
\begin{tabular}{c}
supervised subspace\\
learning
\end{tabular}
&
\begin{tabular}{c}
fast
\end{tabular}
&
low
&
strong
&
weak
\\ [0 ex] \hline

9&
\begin{tabular}{c}
Codebook learning\\
based appearance models
\end{tabular}
&
\begin{tabular}{c}
Vector quantization
\end{tabular}
&
\begin{tabular}{c}
slow
\end{tabular}
&
high
&
strong
&
strong
\\ [0 ex] \hline

\end{tabular}
}
\end{table*}

\section{Benchmark resources for visual object tracking} \label{sec:benchmark}

To evaluate the performance of various tracking algorithms,
one needs the same test video dataset, the ground truth, and the implementation of the competing tracking algorithms.
Tab.~\ref{tab:tracking_datasets} lists the current major resources available to the public.

Another important issue is how to evaluate tracking algorithms in a qualitative or quantitative manner.
Typically, qualitative evaluation is based on intuitive perception by human.
Namely, if the calculated target regions cover more true object regions and
contain fewer non-object pixels, the tracking algorithms
are considered to achieve better tracking performances; otherwise, the tracking algorithms
perform worse. For a clear illustration, a qualitative comparison of several representative
visual representations is provided in
Tab.~\ref{tab:feature_evaluations} in terms of computational speed
as well as handling
occlusion, illumination variation, and
shape deformation capabilities.
Moreover, Tab.~\ref{tab:model_evaluations} provides a qualitative
comparison of several representative statistical modeling-based
appearance models in terms of computational speed,
memory usage, online adaptability, and discriminability.

In contrast, a quantitative evaluation relies heavily on the
ground truth annotation. If objects of interest are annotated
with bounding boxes, a quantitative evaluation
is performed by computing the positional errors of four corners
between the tracked bounding boxes and the ground truth.
Alternatively, the overlapping ratio
between the tracked bounding boxes (or ellipses) and the ground truth can be calculated
for the quantitative evaluation: $r = \frac{A_{t}\bigcap A_{g}}{A_{t}\bigcup A_{g}}$,
where $A_{t}$ is the tracked bounding box (or ellipse) and $A_{g}$ is the ground truth.
The task of ground truth annotation with bounding boxes
or ellipses is difficult and time-consuming. Consequently, researchers
take a point-based annotation strategy for the quantitative evaluation.
Specifically, they either record object center locations as the ground truth
for simplicity and efficiency, or mark several points within the object
regions by hand as the ground truth for accuracy (e.g., seven mark points are used in the dudek face sequence~\cite{Limy-Ross17}).
This way, we can compute the positional residuals between
the tracking results and the ground truth for the quantitative evaluation.

\section{Conclusion and future directions}
\label{sec:Conclusion}

    In this work, we have presented a survey of 2D appearance models for visual object tracking.
    The presented survey takes a
    module-based organization to review the literature of two important modules in 2D appearance models:
    visual representations and
    statistical modeling schemes for tracking-by-detection, as shown in
    Fig.~\ref{fig:architecture}. The visual representations focus more on how to robustly describe
    the spatio-temporal characteristics of object appearance, while the
    statistical modeling schemes for tracking-by-detection put
    more emphasis on how to capture the generative/discriminative statistical information of the object regions.
    These two modules are closely related and interleaved with each other.
    In practice, powerful appearance models depend on not only effective visual representations but also robust statistical models.

    In spite of a great progress in 2D appearance models in recent years, there are still several issues remaining
    to be addressed:

\begin{itemize}

\item Balance between tracking robustness and tracking accuracy.
Existing appearance models are
incapable of simultaneously guaranteeing tracking robustness and  tracking accuracy.
To improve the tracking accuracy, more visual features and geometric constraints are incorporated into the appearance models, resulting in a precise object localization in the situations of particular appearance variations. However, these
visual features and geometric constraints can also lower the generalization capabilities of the appearance models
in the aspect of undergoing other appearance variations.
On the other hand, to improve the tracking robustness, the appearance models relax some constraints on a precise
object localization, and thus allow for more ambiguity of the object localization.
Thus, balancing
tracking robustness and tracking accuracy is an interesting research topic.

\item Balance between simple and robust visual features. In computer vision, designing both simple and robust
visual features is one of the most fundamental and important problems.
In general, simple visual features have a small number of components. As a result,
they are
computationally efficient, but have a low
discriminability. In contrast, robust visual features often have a large number of components. Consequently, they are computationally
expensive, and have sophisticated parameter settings.
Therefore, how to keep a good balance between simplicity and robustness
plays an important role in visual object tracking.

\item 2D and 3D information fusion.
2D appearance models are computationally efficient and simple to implement. Due to the
information loss of 3D-to-2D projections,  2D appearance models cannot accurately estimate the poses of the tracked objects, leading to the sensitivity to occlusion and out-of-plane
rotation. In contrast,  3D appearance models are capable of precisely characterizing the 3D pose
of the tracked objects, resulting in the robustness to occlusion and out-of-plane rotation.
However,  3D appearance models require a large parameter-search space for 3D pose estimation,
resulting in expensive computational costs. Therefore, combining the advantages of  2D and
3D appearance models is a challenging research topic. To accelerate the pose estimation process of
the 3D appearance models, a possible solution is to use the tracking results of the 2D appearance
models as the initialization of the 3D appearance models. However, how to effectively transfer from 2D
tracking to 3D tracking is still an unsolved problem.

\item Intelligent vision models.
Inspired by the biological vision, a number of high-level salient region features are proposed to capture
the salient semantic information of an input image. These salient region features
are relatively stable during the process of tracking, while they rely heavily on
salient region detection which may be affected by noise or drastic illumination variation.
Unreliable saliency detection leads to
many feature mismatches across frames.
Consequently, it is necessary to build an intelligent vision model that can
robustly track these salient region features across frames like what human vision offers.

\item Camera network tracking.
Typically, the appearance models are based on a single camera, which provides a very limited
visual information of the tracked objects. In recent years, several appearance models using multiple overlapping
cameras are proposed to fuse different visual information from different viewpoints.
These appearance models usually deal with the problem of object tracking in the same scene
monitored by different cameras.  Often they cannot complete the tracking task of the same object in different
but adjacent scenes independently. In this case, tracking in a large camera network needs to be established for a long-term
monitoring of the objects of interest. However, how to transfer the target information from one camera sub-network to another
is a crucial issue that remains to be solved.

\item Low-frame-rate tracking. Due to the hardware limits of processing speed
and memory usage, mobile devices or micro embedded systems usually produces the video data with a low frame rate
(e.g., abrupt object motion), which makes the tracking job challenging.
In this situation, the appearance models needs to have a
good generalization and adaptation capability of online coping with the object
appearance variations during tracking. Therefore, it is crucial to construct
a robust appearance model with efficient visual modeling and effective statistical modeling
for real-time applications.
\end{itemize}

\begin{spacing}{1.25}
{
\bibliographystyle{acmtrans}

}
\end{spacing}

\end{document}